\documentclass{article}

\usepackage[preprint]{neurips_2026}

\usepackage[utf8]{inputenc} 
\usepackage[T1]{fontenc}    
\usepackage{hyperref}       
\usepackage{url}            
\usepackage{booktabs}       
\usepackage{amsfonts}       
\usepackage{nicefrac}       
\usepackage{microtype}      
\usepackage{xcolor}         
\usepackage[dvipsnames]{xcolor}
\usepackage{multicol}
\usepackage{multirow}
\usepackage{wrapfig}
\usepackage{subfigure}
\usepackage{subcaption}
\usepackage{amsmath}
\usepackage{pifont}                         
\usepackage{graphicx}
\usepackage{fancyhdr,amssymb}
\usepackage[ruled,vlined]{algorithm2e}
\usepackage{circledsteps}
\usepackage{paralist}
\newcommand{\gmark}
{\textcolor{ForestGreen}{\ding{51}}}  
\newcommand{\xmark}{\textcolor{BrickRed}{\ding{55}}}
\definecolor{GainBright}{RGB}{0,114,178}   
\newcommand{\gain}[2]{\textbf{#1}{\scriptsize\,(\textcolor{GainBright}{#2})}}
\newcommand{\nogain}[2]{{#1}{\scriptsize\,(\textcolor{GainBright}{#2})}}
\usepackage{colortbl}
\usepackage{makecell}

\usepackage{minitoc}

\usepackage{amssymb,amsmath,amsthm}
\usepackage{pgf}
\usepackage{pgfplots}

\usepackage{transparent}

\def\our{REMIX} 

\title{Stop Marginalizing My Dreams: Model Inversion via Laplace Kernel for Continual Learning}

\author{%
  {\bf Patryk Krukowski} \\
  Jagiellonian University \\
  \texttt{patryk.krukowski@doctoral.uj.edu.pl} \\
  \and
  {\bf Jacek Tabor} \\
  Jagiellonian University \\
  \and
  {\bf Przemys{\l}aw Spurek} \\
  Jagiellonian University \\
  IDEAS Research Institute\\
  \and
  {\bf Marek \'Smieja} \\
  Jagiellonian University \\
  \and
  {\bf {\L}ukasz Struski} \\
  Jagiellonian University \\
}

\begin{document}

\maketitle

\begin{abstract}
Data-free continual learning (DFCIL) relies on model inversion to synthesize pseudo-samples and mitigate catastrophic forgetting. However, existing inversion methods are fundamentally limited by a simplifying assumption: they model feature distributions using diagonal covariance, effectively ignoring correlations that define the geometry of learned representations. As a result, synthesized samples often lack fidelity, limiting knowledge retention. In this work, we show that modeling feature dependencies is a key ingredient for effective DFCIL. We introduce \our{}, a structured covariance modeling framework that enables scalable full-covariance modeling without the prohibitive cost of dense matrix inversion and log-determinant computation. By leveraging a Laplace kernel parameterization, \our{} captures structured feature dependencies using memory that scales linearly with the feature dimensionality, while requiring only an additional logarithmic factor in computation. Modeling these correlations produces more coherent synthetic samples and consistently improves performance across standard DFCIL benchmarks. Our results demonstrate that moving beyond diagonal assumptions is essential for effective and scalable data-free continual learning. Our code is available at \url{https://github.com/pkrukowski1/REMIX-Model-Inversion-via-Laplace-Kernel}.
\end{abstract}

\section{Introduction}

Continual learning \cite{mccloskey1989catastrophic,hsu2018re} represents a fundamental challenge in artificial intelligence because deep neural networks typically suffer from catastrophic forgetting when learning tasks sequentially. In many practical scenarios, such as medical or financial applications, the original training data cannot be stored due to strict privacy regulations or memory limits \cite{kaissis2020secure,de2021continual,nayak2019zero}. This has led to the emergence of data-free class-incremental learning, in which model inversion techniques are employed to generate synthetic samples that serve as proxies for forgotten knowledge. 

\begin{figure}[thbp]
    \centering
    \def\svgwidth{\linewidth}
    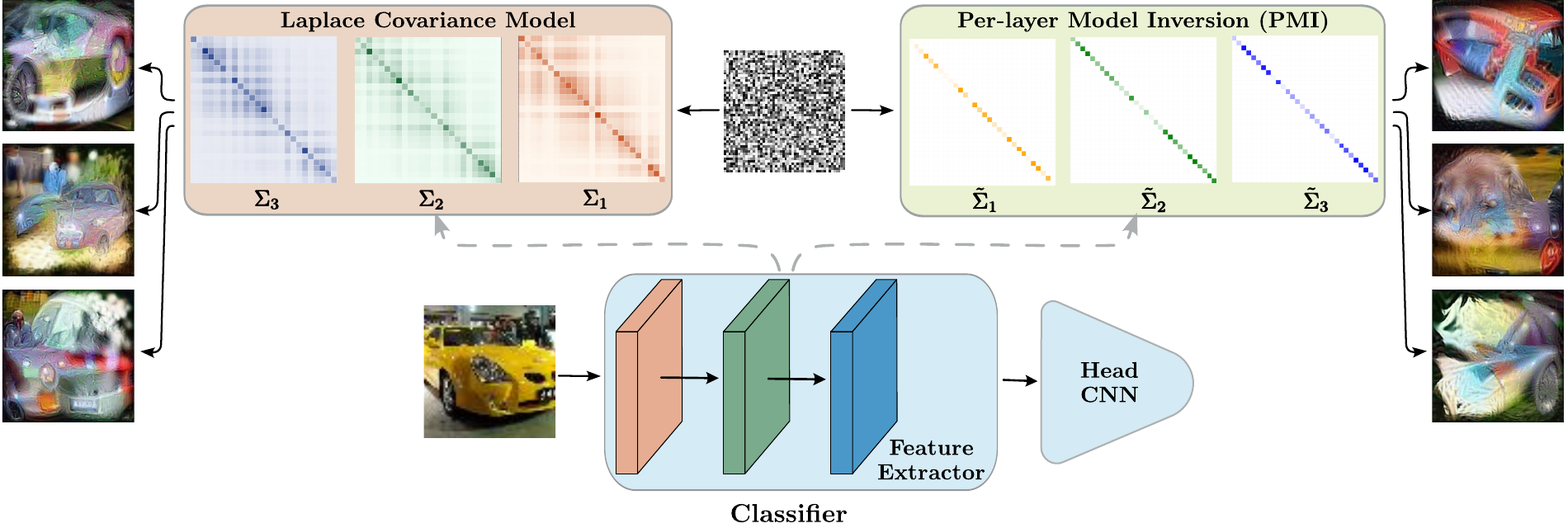
    \caption{Overview of the shared feature extraction pipeline in DFCIL used by the proposed \textit{Laplace Covariance Model (LCM)} and the \textit{Per-layer Model Inversion (PMI)}. 
    While PMI relies on a simplified diagonal approximation of covariance that assumes independence and ignores cross-feature correlations, our
    LCM captures full covariance structure, modeling spatial dependencies.}
    \label{fig:method}
\end{figure}

The foundation of this field was established by DeepInversion~\cite{yin2020dreaming}, which demonstrated that semantically meaningful images can be recovered by optimizing the statistics stored within the batch normalization layers of a frozen model. This mechanism was successfully adapted for continual learning in ABD~\cite{smith2021always}, where the process of constantly dreaming about previous classes serves as a substitute for a physical memory buffer. Further improvements were introduced by R-DFCIL~\cite{gao2022r}, which focused on preserving topological relations and geometric structures in the feature space for more stable knowledge transfer. The current state of the art in this trajectory is PMI \cite{tongmodel}, which utilizes precise layer-specific modeling of feature map statistics. 

Despite these significant advancements, all existing methods are constrained by a critical limitation stemming from the assumption of strictly diagonal covariance matrices. This severe oversimplification completely marginalizes the spatial and inter-channel correlations that are naturally present in deep representations. As a result, the data synthesis process is inherently biased, and the recovered samples fail to capture the full complexity of the original distributions. 

In this work, we introduce \our{} (\textbf{RE}constructing \textbf{M}ultivariate \textbf{I}nteractions via e\textbf{X}ponential kernel) as a transformative solution to this problem by proposing a novel covariance estimator, which represents our primary contribution. Our method redefines the model inversion process by leveraging the mathematical properties of the Laplace kernel to model local dependencies explicitly. 
By employing a structured parameterization, we obtain a tridiagonal precision matrix that captures essential correlations without the overhead of dense covariance matrices. The resulting representation requires memory that scales linearly with the feature dimensionality and enables efficient computation of likelihoods, log-determinants, and inverse operations with only an additional logarithmic factor in runtime. Empirical results show that structured covariance modeling produces more coherent synthesized samples and improves knowledge retention in continual learning. An overview of our method is shown in Figure~\ref{fig:method}.


The main contributions of this paper are summarized as follows:
\begin{compactitem}
    \item We identify the lack of structured correlation modeling as a key limitation of existing DFCIL methods, leading to less effective model inversion and weaker knowledge retention.

    \item We introduce a structured covariance estimator based on the exponential Laplace kernel, which captures feature dependencies through a tridiagonal precision matrix while requiring only $3N$ parameters, where $N$ denotes the feature dimensionality.

    \item We propose the \our{} framework, which integrates structured covariance modeling into the inversion process and achieves state-of-the-art performance on standard DFCIL benchmarks while retaining linear memory complexity and efficient log-linear likelihood evaluation.
\end{compactitem}

\section{Related Work}

In this section, we review the existing literature that forms the foundation of our research. We first explore general approaches to class-incremental learning, in which models are updated sequentially and often rely on physical rehearsal buffers or explicit modeling of feature distributions. Subsequently, we delve into data-free class-incremental learning, which is the primary focus of our work. Within this domain, we trace the evolution from early generative replay methods to modern model inversion techniques, highlighting the critical structural limitations that our proposed covariance estimator aims to resolve.

\textbf{Class Incremental Learning (CIL).}
Class-incremental learning aims to learn new categories continuously from a data stream while maintaining performance on old classes. The main challenge in this domain is catastrophic forgetting. To alleviate this issue, most traditional methods rely on rehearsal strategies that store a small subset of historical data in a memory buffer. For instance, iCaRL~\cite{rebuffi2017icarl} combines a nearest mean classifier with knowledge distillation to retain past knowledge. Following this direction, various approaches have been proposed to improve representation learning and preserve class geometry, such as spatial distillation in PODNet~\cite{douillard2020podnet} and other structural constraints \cite{castro2018end, hou2019learning, yu2020semantic}. Memory management and data sampling strategies are also extensively studied to maximize the utility of the limited buffer \cite{bang2021rainbow, prabhu2020gdumb, liu2021adaptive}. Recently, another promising line of research focuses on freezing feature extractors and modeling feature distributions. Methods like FeCAM~\cite{goswami2023fecam}, FeTrIL~\cite{petit2023fetril}, and AdaGauss~\cite{rypesc2024task} estimate covariance matrices or feature prototypes to stabilize the continuously evolving classifier without explicitly rehearsing past images.

\textbf{Data Free Class Incremental Learning (DFCIL).}
In privacy-sensitive applications, storing raw data is strictly prohibited, thereby prompting the emergence of data-free class-incremental learning. Early attempts to solve this without data rehearsal relied heavily on knowledge distillation directly applied to the new data, such as LwF \cite{li2017learning}. To compensate for missing past data, researchers explored generative replay, in which a generative model is trained alongside the classifier to synthesize past exemplars \cite{shin2017continual,cong2020gan,kemker2018fearnet,wu2018memory,ye2020learning}. These approaches are often inspired by brain memory consolidation \cite{van2020brain}, and their theoretical foundations have been well established \cite{nagarajan2018theoretical}. However, training complex generators sequentially is notoriously unstable and computationally expensive. To bypass this bottleneck, modern data-free methods invert the frozen classifier itself. DeepInversion~\cite{yin2020dreaming} pioneered the synthesis of semantically meaningful images by optimizing random noise to match the batch normalization statistics of a pretrained model. This inversion technique was directly adopted for continual learning in ABD~\cite{smith2021always} to prevent forgetting via dreaming. R-DFCIL~\cite{gao2022r} further improved this framework by introducing relation-guided knowledge distillation, ensuring that the geometric structure of the feature space is preserved during sequential learning. More recently, PMI \cite{tongmodel} introduced layer-specific modeling and alignment to more accurately capture internal distributions. Despite these incredible advances, all these inversion-based techniques fundamentally rely on diagonal covariance assumptions, setting the stage for our proposed method, which explicitly captures structural correlations.






\section{Method}\label{sec:method}



In this section, we introduce \textbf{\our{}}, a model inversion framework for data-free class-incremental learning (DFCIL), which \textbf{RE}constructs \textbf{M}ultivariate \textbf{I}nteractions via e\textbf{X}ponential kernel. The key idea is to move beyond independent feature matching and explicitly model structured correlations within feature representations during sample synthesis. We achieve this by parameterizing feature covariances with a structured, yet efficient formulation based on the Laplace kernel.

We begin by revisiting the DFCIL setting and identifying a core limitation of existing approaches. We then present our Laplace Covariance Model (LCM) and show how it enables efficient and expressive correlation modeling. Finally, we describe how \our{} integrates LCM into continual learning.

\paragraph{Background and Problem Setup.}

We consider the standard data-free class-incremental learning setting with a sequence of tasks $\{\mathcal{D}_t\}_{t=1}^T$, where each dataset $\mathcal{D}_t = \{(\mathbf{x}_i, y_i)\}_{i=1}^{N_t}$ introduces disjoint label spaces. During training on task $t$, only $\mathcal{D}_t$ is accessible, while past data must not be stored.

To mitigate catastrophic forgetting, prior methods rely on \emph{model inversion} to synthesize pseudo-samples from a frozen classifier. 
This is typically done by matching the joint distribution of intermediate activations of real and synthetic inputs conditioned on a target label $y$. By denoting the activations at the layer $l$ as $\mathbf{o}_l$, with $\mathbf{o}_0 = \mathbf{x}$, the objective is
\begin{equation}
    \min_{p_s(\mathbf{x})} D_{\mathrm{KL}}\big(p_s(\mathbf{o}_L, \dots, \mathbf{o}_0 \mid y) \,\|\, p_r(\mathbf{o}_L, \dots, \mathbf{o}_0 \mid y)\big),
\end{equation}
where $p_s$ and $p_r$ denote the distributions of synthetic and real samples, respectively.

Since the above objective is intractable in practice, existing methods approximate it by matching layer-wise Gaussian statistics~\cite{tongmodel}:
\begin{equation}\label{eq:loss_stat}
    \mathcal{L}_{\mathrm{stat}}^{(l)} =
    D_{\mathrm{KL}}\big(\mathcal{N}(\boldsymbol{\hat{\mu}}_{l}, \hat{\mathbf{\Sigma}}_{l})
    \,\|\, \mathcal{N}(\boldsymbol\mu_{l}, \mathbf{\Sigma}_{l})\big),
\end{equation}
where $(\boldsymbol{\mu}_l, \mathbf{\Sigma}_l)$ and $(\boldsymbol{\hat{\mu}}_l, \hat{\mathbf{\Sigma}}_l)$ denote the statistics of real and synthetic features, respectively\footnote{In practice, inversion objectives additionally incorporate standard auxiliary terms, including Cross-Entropy (CE) supervision for class conditioning and layer-wise MSE losses for feature distillation.}. These components primarily act as regularizers, whereas the fundamental representational bottleneck lies in $\mathcal{L}_{\mathrm{stat}}$, which governs how accurately the synthesized distribution matches the structure of the real feature distribution. Accordingly, our theoretical analysis focuses exclusively on this statistical matching term, while the complete optimization objective is provided in Appendix~\ref{appendix:sec_generator_training_details}.

Modeling full covariance matrices requires storing all pairwise feature interactions and performing expensive matrix inversion and log-determinant operations during likelihood evaluation. In high-dimensional feature spaces, these costs become prohibitive, leading existing methods to adopt diagonal approximations of $\mathbf{\Sigma}_l$ that capture only marginal BatchNorm statistics. Concretely, synthetic inputs $\hat{\mathbf{x}}$ are optimized such that their activations $\hat{\mathbf{h}}_l \in \mathbb{R}^C$ match stored means $\boldsymbol{\mu}_l$ and variances $\boldsymbol{\sigma}_l^2$ at layer $l$. This implicitly assumes independence between feature dimensions, ignoring correlations that are essential for modeling the structure of deep representations.



\paragraph{Laplace Covariance Model.} \label{sec:laplace}

To overcome the trade-off between expressiveness and efficiency, we employ the \emph{Laplace Covariance Model (LCM)} -- a structured parameterization that captures non-trivial feature dependencies while requiring only near-linear computational complexity and linear memory complexity with respect to the feature dimensionality.

\begin{figure}[t]
    \centering
    \def\svgwidth{.9\linewidth}
    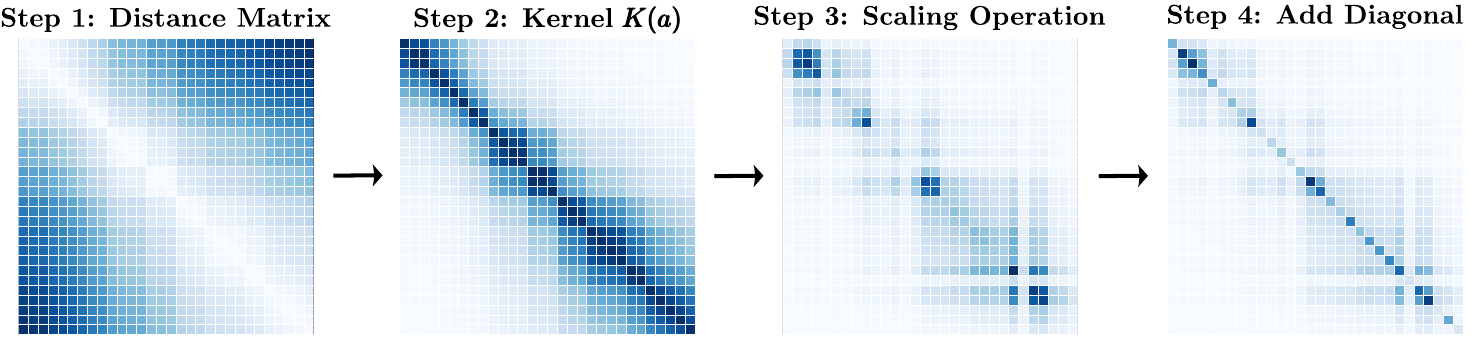
    \caption{Construction of the proposed \our{} covariance matrix $\Sigma \in \mathbb{R}^{C \times C}$ using $\mathcal{O}(C)$ parameters. 
\textbf{(1) Distance Encoding:} Each channel is assigned a learnable coordinate $a_i$, defining pairwise distances $|a_i - a_j|$ that capture channel relationships. 
\textbf{(2) Kernel Mapping:} Distances are converted into correlations via the Laplace kernel $K_{ij} = \exp(-|a_i - a_j|)$, inducing exponential decay and a tridiagonal precision structure. 
\textbf{(3) Scaling:} The kernel is rescaled as $\Sigma' = \mathrm{diag}(w) K \mathrm{diag}(w)$, allowing channel-wise variance modulation. 
\textbf{(4) Diagonal Adjustment:} A diagonal term $d$ is added to obtain $\Sigma = \Sigma' + \mathrm{diag}(d)$, ensuring stability and modeling independent noise.}
    \label{fig:cov_flow}
\end{figure}

For a feature vector of dimension $C$, we associate each dimension $i$ with a scalar latent coordinate $a_i \in \mathbb{R}$. Pairwise correlations are induced via a Laplace kernel,
\begin{equation}
K_{ij}(a) = \exp(-|a_i - a_j|),
\end{equation}
which encodes the intuition that channels closer in the latent space exhibit stronger interactions. To increase expressivity, we introduce a learnable scaling vector $\mathbf{w} \in \mathbb{R}^C$, as well as an independent diagonal component controlling per-dimension variance. The resulting covariance parameterization takes the form:
\begin{equation} \label{eq:cov}
\mathbf{\Sigma} = \mathrm{diag}(\mathbf{d}) + \mathrm{diag}(\mathbf{w})\, K(\mathbf{a})\, \mathrm{diag}(\mathbf{w}),
\end{equation}
with $\mathbf{d} = \mathrm{softplus}(\mathbf{u}) + \varepsilon$, where $\mathbf{u} \in \mathbb{R}^C$ controls the diagonal variance, $\mathrm{diag}(\cdot)$ constructs a diagonal matrix, and $\varepsilon > 0$ ensures numerical stability (see~Fig.~\ref{fig:cov_flow}). This formulation captures structured feature correlations using only a linear number of parameters, unlike dense covariance matrices that scale quadratically with the feature dimension.

Likelihood-based objectives, see Eq.~\eqref{eq:loss_stat}, are a natural choice for Gaussian modeling, but directly optimizing them under structured covariance parameterizations requires computing log-determinants and matrix inverses, which leads to non-trivial computational and numerical overhead. We instead optimize a Frobenius-norm objective, which avoids explicit covariance inversion while matching second-order statistics and reduces computation to element-wise operations and quadratic forms. 

Let $\hat{\mathbf{\Sigma}} = \frac{1}{N} \sum_{i=1}^N \mathbf{v}_i \mathbf{v}_i^\top$ denote the empirical covariance of centered features. Instead of optimizing Eq.~\eqref{eq:loss_stat}, we resort to minimize $\| \mathbf{\Sigma} - \hat{\mathbf{\Sigma}}\|^2_{\mathrm{F}}$,
where $\mathbf{\Sigma}$ denotes the LCM parameterization, see Eq.~\eqref{eq:cov}. Expanding the objective yields:
\begin{equation} \label{eq:frobenius_loss_function}
\| \mathbf{\Sigma} - \hat{\mathbf{\Sigma}}\|^2_{\mathrm{F}} = \underbrace{\|\mathbf{\Sigma}\|_{\mathrm{F}}^2}_{\text{model term}} - \underbrace{\frac{2}{N} \sum_{i=1}^N \mathbf{v}_i^\top \mathbf{\Sigma} \mathbf{v}_i}_{\text{data term}} + \text{ const},
\end{equation}
which decomposes into a model-dependent regularization term and a data-alignment term (see Appendix~\ref{appendix:sec_frobenius_norm_derivation} for a detailed evaluation). Under the LCM structure, the model term simplifies to:
\begin{equation}
\|\mathbf{d}\|_2^2 + 2\, \mathbf{d}^\top \mathbf{w}^2 + (\mathbf{w}^2)^\top K(2\mathbf{a}) (\mathbf{w}^2),
\end{equation}
while the data term becomes:
\begin{equation}
\frac{2}{N} \sum_{i=1}^N \left( \mathbf{d}^\top \mathbf{v}_i^2 + (\mathbf{w} \odot \mathbf{v}_i)^\top K(\mathbf{a}) (\mathbf{w} \odot \mathbf{v}_i) \right).
\end{equation}

Crucially, all computations reduce to element-wise operations and quadratic forms of the type $\mathbf{x}^\top K(\mathbf{a}) \mathbf{x}$. These quadratic forms can be computed in $\mathcal{O}(C\log{C})$ time using cumulative (prefix and suffix) sums, avoiding explicit matrix construction (see Appendix~\ref{appendix:sec_prefix_suffix}). This is enabled by the one-dimensional latent structure of $\mathbf{a}$, which induces an ordering over channels and allows kernel interactions to be computed via directional accumulations rather than dense matrix multiplication.


Although $\mathbf{\Sigma}$ is dense, its inverse $\mathbf{\Sigma}^{-1}$ corresponds to a tridiagonal precision matrix due to the Markovian structure induced by the Laplace kernel (see Appendix~\ref{appendix:sec_gmrf}). As a result:
\begin{compactitem}
    \item quadratic forms $\mathbf{x}^\top \mathbf{\Sigma}^{-1} \mathbf{x}$ can be computed in $\mathcal{O}(C\log{C})$ time,
    \item log-determinants and likelihood terms can be evaluated efficiently without dense matrix operations.
\end{compactitem}
Consequently, model inversion can be performed by directly optimizing the log-likelihood objective while remaining scalable in high-dimensional feature spaces, requiring only linear memory and near-linear computation. Detailed efficiency analyses are provided in Appendix~\ref{appendix:sec_additional_experiments}.

\paragraph{\our{}: Integration of LCM into DFCIL.}

We now present \our{}, which integrates LCM into the DFCIL pipeline. Specifically, we describe: (i) feature distribution modeling across different architectures, (ii) continual aggregation of feature statistics, and (iii) generator-based model inversion.

We decompose the network into a feature extractor $f_{\mathrm{feat}}(\cdot;\mathbf{\theta})$ and a classifier head $f_{\mathrm{head}}(\cdot;\mathbf{\phi})$, and employ a generator $g(\cdot;\mathbf{\psi})$ to synthesize pseudo-samples.

Instead of storing raw data, we maintain compact feature statistics aggregated over previously observed tasks. Prior approaches retain only marginal statistics and therefore assume diagonal covariance structures. In contrast, \our{} additionally stores the LCM parameters $(\mathbf{d}, \mathbf{w}, \mathbf{a})$, enabling structured modeling of feature correlations.

For convolutional networks, BatchNorm layers naturally provide estimates of feature means and variances. We therefore use BatchNorm statistics to preserve marginal feature distributions, while LCM models the correlation structure between feature dimensions. After aggregation across tasks, the resulting covariance representation is re-scaled to remain consistent with the variances stored in BatchNorm statistics.

For architectures without BatchNorm (e.g., ViTs~\cite{dosovitskiy2021imageworth16x16words}), we explicitly record layer-wise activations and employ a sequential layer-wise inversion strategy~\cite{tongmodel,kazemi2024learninvertingclipmodels}.
During training on task $t$, we estimate LCM parameters $(\mathbf{d}_t, \mathbf{w}_t, \mathbf{a}_t)$ for each layer using the Frobenius objective. To aggregate knowledge across tasks, we maintain a running covariance estimate:
\begin{equation}
\mathbf{\Sigma}_{1:t+1} = \mathbf{\Sigma}_{1:t} \cdot \frac{|\mathcal{Y}_{1:t}|}{|\mathcal{Y}_{1:t+1}|}
+ \mathbf{\Sigma}_{t+1} \cdot \frac{|\mathcal{Y}_{t+1}|}{|\mathcal{Y}_{1:t+1}|},
\end{equation}
where $|\mathcal{Y}_{1:t}|$ denotes the number of classes observed up to task $t$, and $|\mathcal{Y}_{t+1}|$ is the number of new classes introduced at task $t+1$. This is followed by re-fitting the LCM parameters to the aggregated covariance, allowing structured feature relationships to be preserved across tasks without retaining raw samples.

In model inversion stage, pseudo-samples are generated using the generator $g(z;\mathbf{\psi})$ by enforcing both semantic and statistical constraints. Unlike prior methods that match only marginal statistics, we impose a full-covariance objective:
$\log p(\mathbf{h}) \propto - (\mathbf{h} - \boldsymbol{\mu})^\top \mathbf{\Sigma}^{-1} (\mathbf{h} - \boldsymbol{\mu})$,
where $(\boldsymbol{\mu}, \mathbf{\Sigma})$ are defined by the aggregated LCM parameters. Due to the tridiagonal structure of the precision matrix $\mathbf{\Sigma}^{-1}$, these likelihood terms can be evaluated in $\mathcal{O}(C\log{C})$ time, enabling efficient model inversion despite modeling full covariance (see Appendix~\ref{appendix:sec_gmrf}).

The likelihood objective is combined with standard inversion losses (e.g., classification alignment and image priors) to guide the generator toward realistic and diverse samples. The full generator optimization objective and detailed description of the CFS procedure are provided in Appendix~\ref{appendix:sec_generator_training_details}.

We follow a standard two-stage DFCIL procedure to train \our{}. First, the feature extractor and classifier are trained using real data from the current task and generated samples from previous tasks, with distillation to mitigate forgetting. Then, the classifier is fine-tuned on a balanced mixture of real and synthetic data. Formal definitions of all loss terms and implementation details, including loss scaling, are provided in Appendix~\ref{appendix:sec_feature_extractor_and_classifier_training_details}.

\section{Experiments}\label{sec:experiments}
In this section, we describe the experimental setup used to evaluate \our{}, including the considered benchmarks and datasets. We then present quantitative results demonstrating improvements over SOTA methods, followed by ablation studies.

\paragraph{Experimental Setup.}

We evaluate \our{} using two standard continual learning metrics: \textit{average incremental accuracy} and \textit{last task accuracy} (see Appendix~\ref{appendix:sec_metrics_definition} for formal definitions).

Experiments are conducted on CIFAR-100~\cite{krizhevsky2009learning}, Tiny-ImageNet~\cite{Wu2017TinyIC}, and CUB-200~\cite{cub}, following the PMI~\cite{tongmodel} and R-DFCIL~\cite{gao2022r} protocols with identical task splits and evaluation settings. CIFAR-100 and Tiny-ImageNet are evaluated with a ResNet-32 backbone, while CIFAR-100 and CUB-200 are additionally considered in a CLIP-based setting with a frozen image encoder~\cite{pmlr-v139-radford21a}. All experiments are performed in the data-free regime, where no past data are retained and pseudo-samples are generated via model inversion. We compare \our{} against prior DFCIL methods as well as selected approaches that rely on real data buffers. Additional details on task configurations and training setups are provided in Appendix~\ref{appendix:sec_implementation_details}.

To ensure a fair comparison, we adopt all hyperparameters from PMI, except for the Frobenius regularization weight $\lambda_{\mathrm{F}}$, which is selected via grid search. The full search space and implementation details are given in Appendix~\ref{appendix:sec_implementation_details}.





\paragraph{Results on ResNet-32 Backbone.}

Table~\ref{tab:resnet_cl} reports the average incremental accuracy on CIFAR-100 and Tiny-ImageNet using a ResNet-32 backbone under different task splits. Overall, \our{} consistently outperforms prior data-free approaches across all settings. Compared to the strongest baseline, PMI, our method achieves systematic improvements on both datasets, with gains becoming more pronounced as the number of tasks increases. 

On CIFAR-100, \our{} improves performance across all splits, with the largest gain of \textbf{+1.48} percentage points (p.p.) observed in the 10-task setting. Similarly, on Tiny-ImageNet, \our{} achieves consistent improvements, reaching up to \textbf{+1.44} p.p. in the most challenging 20-task scenario. These results suggest that explicitly modeling feature correlations during inversion produces more coherent pseudo-samples, thereby reducing error accumulation and mitigating catastrophic forgetting over long task sequences.

\begin{table}[htbp!]
\caption{Average incremental accuracy (\%) on CIFAR-100 and Tiny-ImageNet using a ResNet-32 backbone. Results are averaged over five runs. Baseline results are taken from~\cite{tongmodel}. Values in parentheses indicate the improvement of \our{} over the best baseline.}
\label{tab:resnet_cl}
\centering
\small
\setlength{\tabcolsep}{2.4pt}
\begin{tabular}{@{}l c c c c c c c@{}}
\toprule
\multirow{2}{*}{Method} & \multirow{2}{*}{\parbox[c]{2cm}{\centering Model\\Inversion}}  & \multicolumn{3}{c}{CIFAR-100} & \multicolumn{3}{c}{Tiny-ImageNet}   \\
\cmidrule(lr){3-5}\cmidrule(lr){6-8}
                 &        & 5 task      & 10 task     & 20 task    & 5 task     & 10 task    & 20 task \\
\midrule
Upper bound      & \xmark & 70.59       & 70.59       & 70.59      & 55.25      & 55.25      & 55.25      \\
DeepInversion    & \gmark & 20.48       & 11.26       & 5.63       & -          & -          & -          \\
ABD              & \gmark & 48.84       & 36.75       & 24.40      & 30.83      & 23.17      & 14.61      \\
R-DFCIL          & \gmark & 49.87       & 41.80       & 31.54      & 35.33      & 29.05      & 24.85      \\
PMI w/o CFS & \gmark & 52.05 & 43.23 & 32.23 & 37.65 & 32.09 & 25.51 \\
PMI         & \gmark & 52.38 & 43.90 & 32.60 & 37.90 & 32.43 & 25.67 \\
\midrule
\rowcolor{Orange!10}
\textbf{\our{}}         & \gmark & \gain{52.94}{+0.56} & \gain{45.38}{+1.48} & \gain{33.11}{+0.51} & \gain{38.64}{+0.74} & \gain{33.39}{+0.96} & \gain{27.11}{+1.44} \\
\bottomrule
\end{tabular}
\vspace{-4pt}
\end{table}

Importantly, the improvements are obtained without modifying the underlying training protocol or hyperparameters of PMI, demonstrating that the gains stem directly from the proposed covariance modeling rather than additional tuning.

\paragraph{Results on ViT Backbone.}

Table~\ref{tab:clip_cl} presents continual learning performance using a ViT-based CLIP backbone on CIFAR-100 and CUB-200. We report both average and last task accuracy to capture overall performance and final retention. 

Overall, \our{} combined with MoE-Adapter achieves the best results among data-free approaches and remains competitive with methods that rely on real image buffers. Compared to the strongest baseline (PMI + MoE-Adapter), our method consistently improves performance across both datasets and metrics. On CIFAR-100, \our{} achieves gains of \textbf{+0.12} p.p. in average accuracy and \textbf{+0.31} p.p. in last-task accuracy. On the more challenging fine-grained CUB-200 benchmark, \our{} remains competitive in average accuracy (-0.02) while achieving a clear improvement of \textbf{+0.43} in final accuracy, indicating stronger retention of previously learned classes.

\begin{table}[t]
\centering
\caption{Average incremental accuracy (\%) on CIFAR-100 and CUB-200 using a ViT backbone. Baseline results are taken from~\cite{tongmodel}. Values in parentheses indicate the improvement of \our{} over the best baseline.}
\label{tab:clip_cl}
\small
\setlength{\tabcolsep}{4pt}
\begin{tabular}{lcccccc}
\toprule
\multirow{2}{*}{Method} &
\multirow{2}{*}{\parbox[c]{2cm}{\centering No Real Image\\Buffer}} &
\multicolumn{2}{c}{CIFAR-100} &
\multicolumn{2}{c}{CUB-200} \\
\cmidrule(lr){3-4}\cmidrule(lr){5-6}
 & & Avg. & Last & Avg. & Last \\
\midrule
iCaRL                  & \xmark & 82.20 & 64.15 & 82.39 & 75.11 \\
MEMO                   & \xmark & 85.31 & 75.24 & 77.72 & 65.95 \\
PROOF                  & \xmark & 86.95 & 79.32 & 85.13 & 79.69 \\
CLAP4CLIP              & \xmark & 86.13 & 78.21 & 86.93 & 81.64 \\
\midrule
ZS-CLIP                & \gmark & 76.71 & 66.25 & 66.81 & 53.52 \\
AttriCLIP              & \gmark & 79.31 & 68.45 & 65.26 & 52.12 \\
VPT                    & \gmark & 83.98 & 74.34 & 68.22 & 55.51 \\
CODA-P                 & \gmark & 83.64 & 75.80 & 73.05 & 61.83 \\
MoE-Adapter            & \gmark & 87.29 & 79.40 & 73.34 & 61.07 \\
\midrule
PMI + VPT          & \gmark &
    85.45 & 76.91 &
    70.14 & 57.04 \\
PMI + CODA-P       & \gmark &
    84.78 & 76.23 &
    74.37 & 64.63 \\
PMI + MoE-Adapter  & \gmark &
    88.35 & 81.06 &
    \textbf{78.98} & 67.26 \\
\midrule
\rowcolor{Orange!10}
\textbf{\our{} + MoE-Adapter}  & \gmark &
    \gain{88.47}{+0.12} & \gain{81.37}{+0.31} &
    \nogain{78.96}{-0.02} & \gain{67.69}{+0.43} \\
\bottomrule
\end{tabular}
\end{table}

These results demonstrate that modeling structured feature dependencies during inversion also benefits the CLIP setting. The consistent gains across both datasets indicate improved alignment between synthetic and real feature distributions, resulting in better knowledge retention. Importantly, these improvements are achieved without modifying the underlying CLIP adaptation framework, suggesting that they stem directly from the proposed covariance modeling.

\paragraph{Qualitative Analysis of Generated Samples.}

\begin{figure*}[t]
\centering
\setlength{\tabcolsep}{2pt}

\begin{tabular}{c@{}c@{\quad}c@{\;}c@{}}

& \textbf{Diagonal Covariance} & \textbf{Full-Feature Covariance} \\


\rotatebox{90}{\textbf{\small \phantom{ooo}Car}}
&
\includegraphics[width=0.11\linewidth]{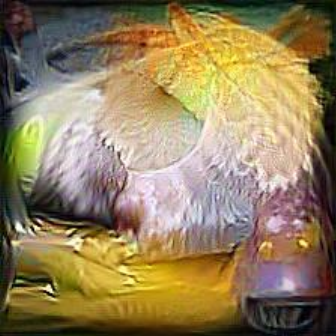}\hspace{1pt}
\includegraphics[width=0.11\linewidth]{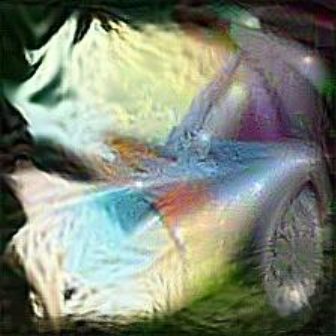}\hspace{1pt}
\includegraphics[width=0.11\linewidth]{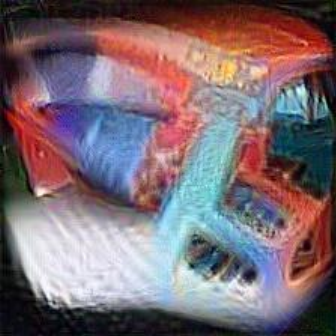}\hspace{1pt}
\includegraphics[width=0.11\linewidth]{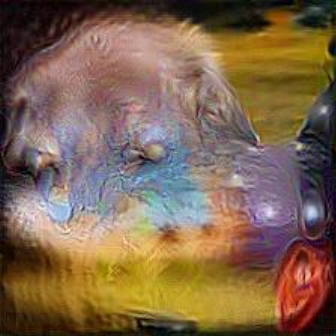}
&
\includegraphics[width=0.11\linewidth]{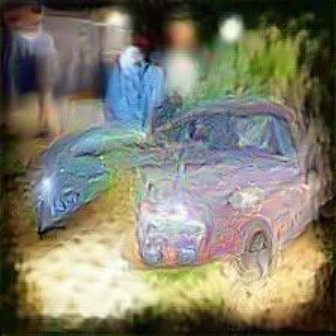}\hspace{1pt}
\includegraphics[width=0.11\linewidth]{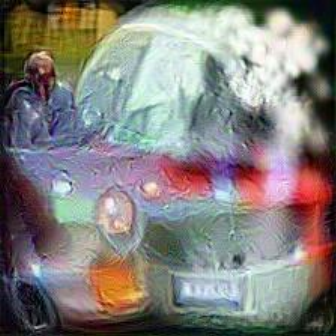}\hspace{1pt}
\includegraphics[width=0.11\linewidth]{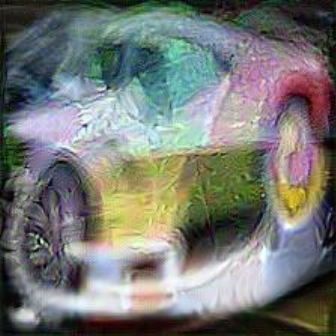}\hspace{1pt}
\includegraphics[width=0.11\linewidth]{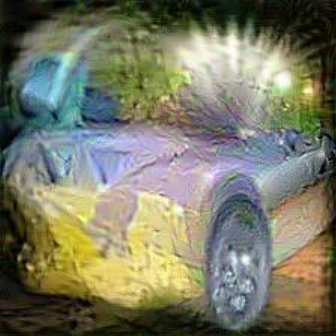} 
&
\multirow{2}{3cm}{\rotatebox{90}{\textbf{\small ResNet-34}}}
\\

\rotatebox{90}{\textbf{\small \phantom{ooo}Dog}}
&
\includegraphics[width=0.11\linewidth]{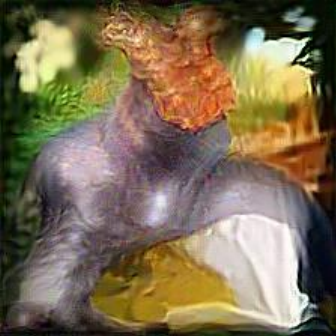}\hspace{1pt}
\includegraphics[width=0.11\linewidth]{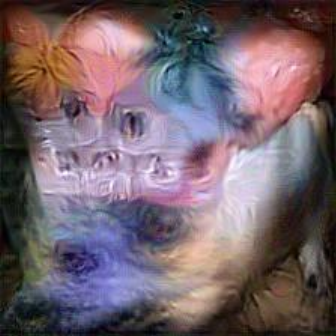}\hspace{1pt}
\includegraphics[width=0.11\linewidth]{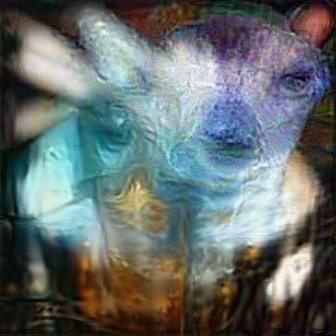}\hspace{1pt}
\includegraphics[width=0.11\linewidth]{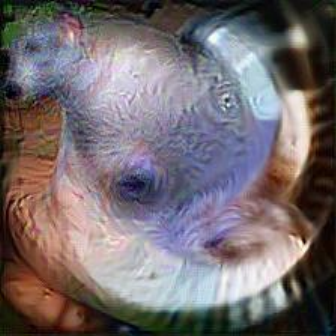}
&
\includegraphics[width=0.11\linewidth]{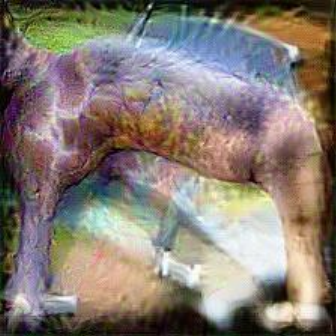}\hspace{1pt}
\includegraphics[width=0.11\linewidth]{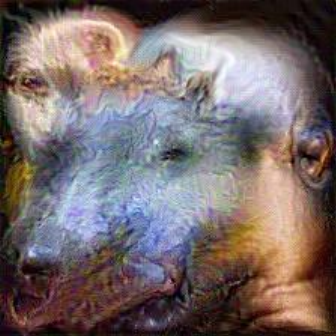}\hspace{1pt}
\includegraphics[width=0.11\linewidth]{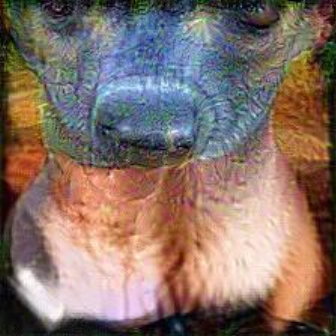}\hspace{1pt}
\includegraphics[width=0.11\linewidth]{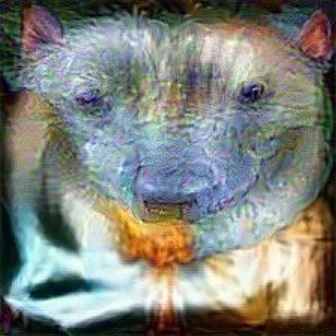}
\\[1em]


\rotatebox{90}{\textbf{\small \phantom{ooo}Car}}
&
\includegraphics[width=0.11\linewidth]{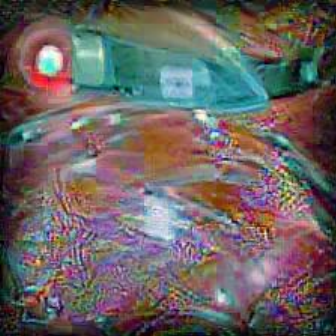}\hspace{1pt}
\includegraphics[width=0.11\linewidth]{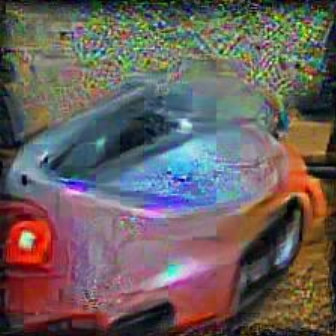}\hspace{1pt}
\includegraphics[width=0.11\linewidth]{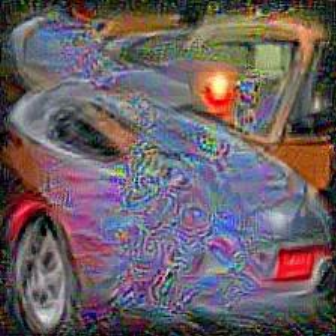}\hspace{1pt}
\includegraphics[width=0.11\linewidth]{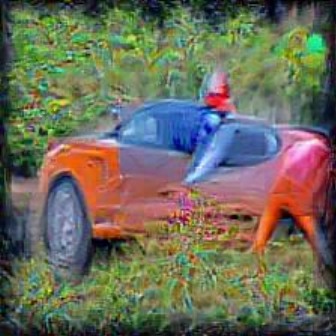}
&
\includegraphics[width=0.11\linewidth]{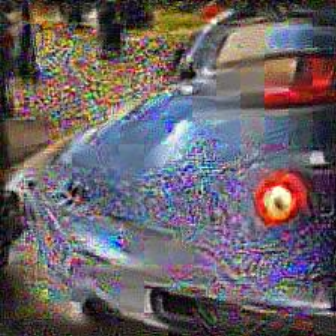}\hspace{1pt}
\includegraphics[width=0.11\linewidth]{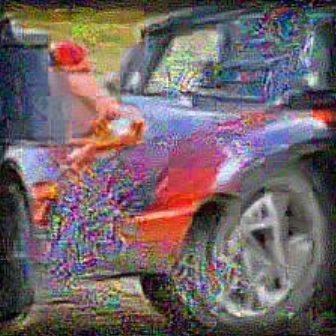}\hspace{1pt}
\includegraphics[width=0.11\linewidth]{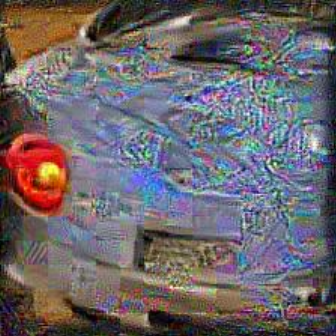}\hspace{1pt}
\includegraphics[width=0.11\linewidth]{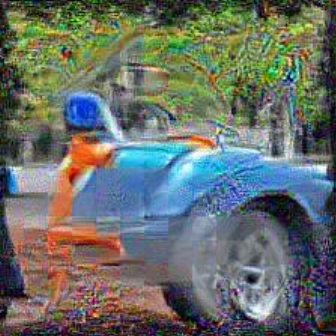}
&
\multirow{2}{3cm}{\rotatebox{90}{\textbf{\small ViT-B/16}}}
\\

\rotatebox{90}{\textbf{\small \phantom{ooo}Dog}}
&
\includegraphics[width=0.11\linewidth]{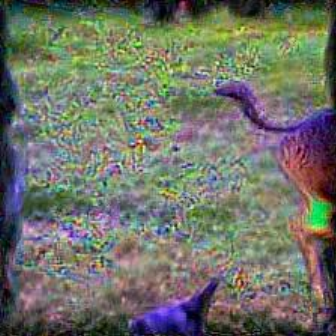}\hspace{1pt}
\includegraphics[width=0.11\linewidth]{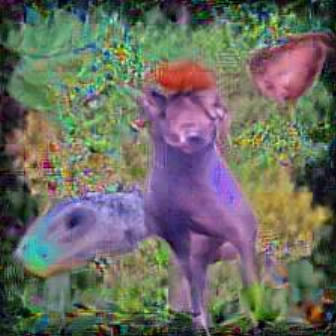}\hspace{1pt}
\includegraphics[width=0.11\linewidth]{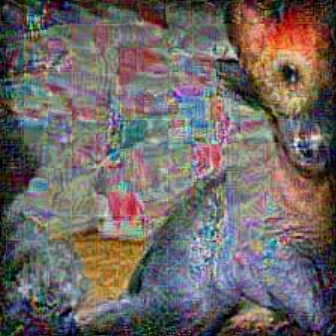}\hspace{1pt}
\includegraphics[width=0.11\linewidth]{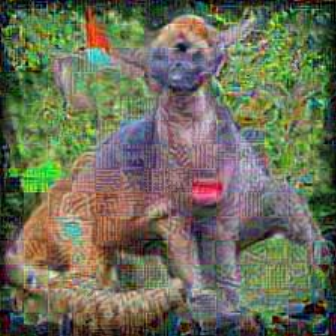}
&
\includegraphics[width=0.11\linewidth]{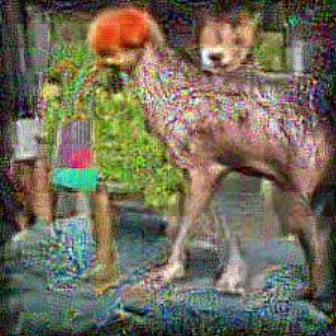}\hspace{1pt}
\includegraphics[width=0.11\linewidth]{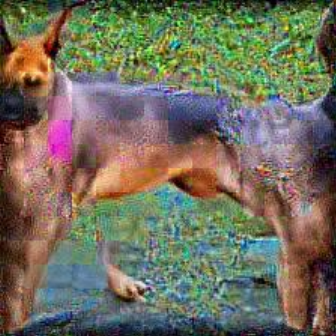}\hspace{1pt}
\includegraphics[width=0.11\linewidth]{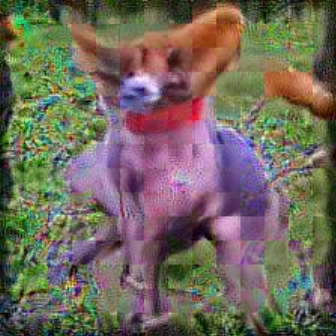}\hspace{1pt}
\includegraphics[width=0.11\linewidth]{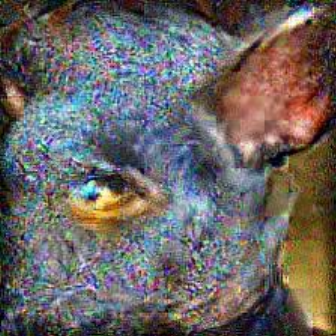}

\end{tabular}

\caption{Qualitative comparison of synthetic samples generated using diagonal covariance modeling and the proposed full-feature covariance modeling. Results are shown for both ResNet-34 and ViT-B/16 backbones. Modeling structured feature dependencies produces samples with improved semantic coherence and more consistent object structure across architectures.}

\label{fig:synthetic_samples}
\end{figure*}

Figure~\ref{fig:synthetic_samples} illustrates the impact of modeling structured feature dependencies during model inversion across both convolutional (ResNet-34) and transformer-based (ViT-B/16) architectures. Methods relying on diagonal covariance assumptions treat feature dimensions independently, often producing samples with artifacts, blurred structures, and inconsistent semantics. In contrast, \our{} captures correlations within feature representations while remaining computationally efficient, even for high-dimensional feature spaces. As a result, the generated samples exhibit clearer object structure, more coherent textures, and stronger semantic consistency across architectures.

\begin{figure}[t]
    \centering
    \resizebox{0.95\linewidth}{!}{\input{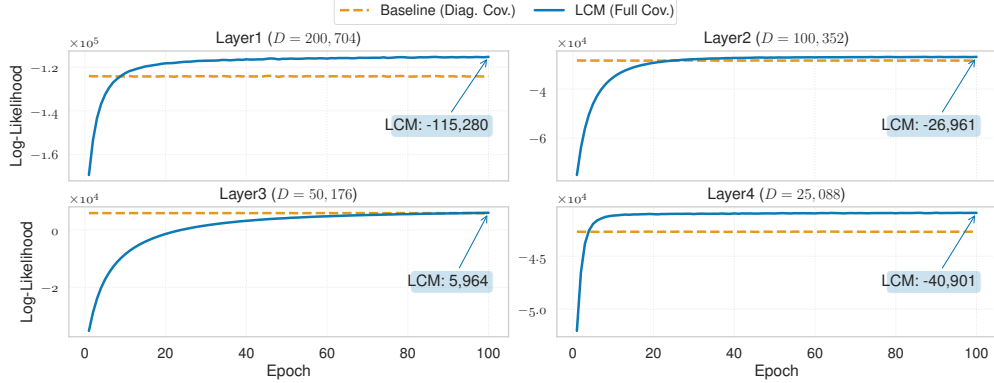}}
    \caption{Log-likelihood comparison between diagonal and full-feature covariance models across ResNet-34 layers (\texttt{layer1}--\texttt{layer4}) on ImageNet classes (dog, car, koala, greenhouse). The proposed LCM (solid line) consistently outperforms the diagonal baseline (dashed line) across all layers.}
    \label{fig:resnet34_ll_analysis}
\end{figure}

To further quantify this effect, we next analyze the log-likelihood of features under the learned distributions (see Figure~\ref{fig:resnet34_ll_analysis}), showing that modeling full feature-map correlations yields a more faithful representation of the underlying feature space. Additional qualitative results for ViT backbones, along with implementation details and a detailed memory analysis, are provided in Appendix~\ref{appendix:sec_additional_experiments}, where we show that \our{} reduces the covariance parameterization cost by over $\mathbf{35{,}000\times}$ compared to dense representations.

\paragraph{Effect of Covariance Modeling.}
We assess the impact of covariance modeling by comparing log-likelihoods under two Gaussian assumptions: diagonal covariance and the proposed LCM. For ResNet-34, we extract intermediate feature representations from \texttt{layer1}--\texttt{layer4} for four ImageNet classes (dog, car, koala, greenhouse), flattening each feature map into a high-dimensional vector ($D = C \!\times\! H \!\times\! W$, up to $\sim 2 \cdot 10^5$).

As shown in Figure~\ref{fig:resnet34_ll_analysis}, LCM consistently achieves higher log-likelihood than the diagonal baseline across all layers. This improvement persists even in extremely high-dimensional regimes, demonstrating that capturing dependencies across the full feature map leads to a more accurate representation of the underlying feature distribution.

Additional results for ViT architectures are provided in Appendix~\ref{appendix:sec_additional_experiments}, where we further examine when full-covariance modeling is beneficial and when diagonal assumptions remain sufficient across transformer layers.

\begin{wrapfigure}{r}{0.5\textwidth}
    \vspace{-1.7\baselineskip}
    \centering
    \resizebox{\linewidth}{!}{\input{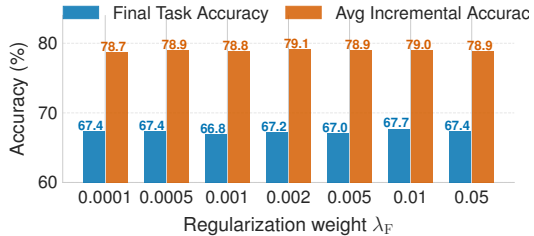}}
    \caption{Sensitivity to $\lambda_{\mathrm{F}}$ on CUB-200 (ViT + MoE-Adapter).}
    \label{fig:vit_lambda_frob_ablation}
    \vspace{-1.0\baselineskip}
\end{wrapfigure}

\paragraph{Sensitivity to $\lambda_{\mathrm{F}}$.}
We analyze the sensitivity of \our{} to the Frobenius regularization weight $\lambda_{\mathrm{F}}$ by sweeping values in the range $[10^{-4}, 5\cdot 10^{-2}]$. The study is conducted on the CUB-200 dataset using a ViT backbone with the MoE-Adapter framework. 

The average incremental accuracy (Figure~\ref{fig:vit_lambda_frob_ablation}, left) remains highly stable across all configurations, varying by less than $0.5\%$, indicating that the overall learning dynamics are robust to the choice of $\lambda_{\mathrm{F}}$. In contrast, the final accuracy $\mathcal{A}_{\text{last}}$ (right) exhibits a clear optimum at $\lambda_{\mathrm{F}} = 0.01$, where the model achieves the best retention performance ($67.69\%$).

    


This behavior reflects a trade-off between structural regularization and plasticity: small values (e.g., $10^{-4}$) provide insufficient constraint, leading to drift in feature structure, while large values (e.g., $0.05$) overly restrict adaptation to new tasks.

\begin{figure}[t]
    \centering
    \resizebox{\linewidth}{!}{\input{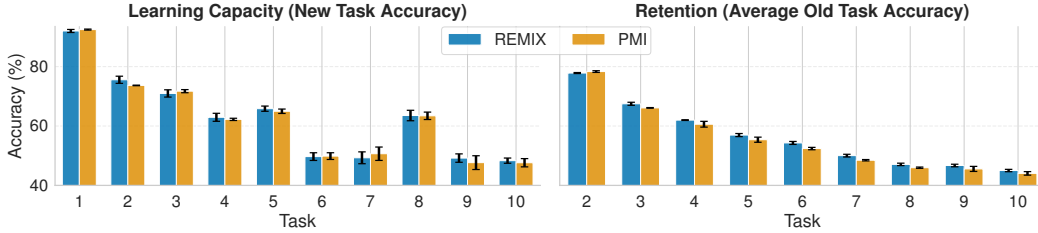}}
    \caption{Task-wise performance comparison on CIFAR-100 (ResNet-32). Solid lines denote mean accuracy over three runs, with shaded regions indicating standard deviation. \our{} consistently outperforms PMI in retaining previously learned knowledge while maintaining similar performance trends across tasks.}
    \label{fig:resnet32_cifar100_acc}
\end{figure}

\textbf{Learning Capacity and Retention.}
We analyze the trade-off between plasticity and stability by comparing \our{} and PMI on CIFAR-100 using a ResNet-32 backbone. Figure~\ref{fig:resnet32_cifar100_acc} reports task-wise performance, where the left plot shows accuracy on the current task (learning capacity), while the right plot shows the average accuracy on previously learned tasks (retention), both averaged over three runs. We observe that \our{} maintains performance comparable to PMI on new tasks, indicating that modeling feature dependencies does not impair learning capacity. At the same time, \our{} consistently achieves higher accuracy on previously learned tasks, demonstrating improved retention. These results indicate that incorporating structured covariance effectively mitigates forgetting while preserving the plasticity required for continual learning. Analogous results on Tiny-ImageNet are provided in Appendix~\ref{appendix:sec_additional_experiments}.

\section{Conclusions}\label{sec:conclusions}

We introduce \our{}, a scalable covariance modeling framework for DFCIL that captures feature correlations using only a linear number of parameters. Integrated into the PMI-based inversion pipeline, \our{} enables efficient full-covariance modeling in high-dimensional feature spaces and improves the reconstruction of intermediate representations beyond diagonal approximations. As a result, the generated pseudo-samples better preserve previously learned knowledge, leading to improved retention in the DFCIL setting.

\paragraph{Limitations.}
The proposed formulation relies on a structured covariance parameterization, which may be less expressive than more flexible generative models. Moreover, since \our{} builds upon the PMI inversion framework, it inherits limitations related to imperfect feature distribution alignment during inversion. Exploring richer distribution modeling and more advanced inversion objectives remains an important direction for future work.


\bibliographystyle{unsrt}



\appendix

\section{Derivation of the Frobenius-Norm Covariance Objective}\label{appendix:sec_frobenius_norm_derivation}

Given $N$ centered feature vectors $\{\mathbf{v}_i\}_{i=1}^N$, the empirical covariance is $\hat{\mathbf{\Sigma}} = \frac{1}{N} \sum_{i=1}^N \mathbf{v}_i \mathbf{v}_i^\top$. We parameterize the model covariance $\mathbf{M}$ using the Laplace kernel $K(\mathbf{a})_{ij} = e^{-|a_i - a_j|}$:
\begin{equation*}
\mathbf{M} = \mathrm{diag}(\mathbf{d}) + \mathrm{diag}(\mathbf{w}) K(\mathbf{a}) \mathrm{diag}(\mathbf{w}).
\end{equation*}

We minimize the squared Frobenius distance $\mathcal{L}_{\mathrm{F}} = \| \mathbf{M} - \hat{\mathbf{\Sigma}} \|_{\mathrm{F}}^2$. Expanding the norm yields:
\begin{equation*}
\| \mathbf{M} - \hat{\mathbf{\Sigma}} \|_{\mathrm{F}}^2 = \langle \mathbf{M}, \mathbf{M} \rangle_{\mathrm{F}} - \frac{2}{N} \sum_{i=1}^N \langle \mathbf{M}, \mathbf{v}_i \mathbf{v}_i^\top \rangle_{\mathrm{F}} + \langle \hat{\mathbf{\Sigma}}, \hat{\mathbf{\Sigma}} \rangle_{\mathrm{F}}.
\end{equation*}
The term $\langle \hat{\mathbf{\Sigma}}, \hat{\mathbf{\Sigma}} \rangle_{\mathrm{F}}$ is constant with respect to the model parameters and is omitted from the optimization objective.

We expand the model self-inner product $\langle \mathbf{M}, \mathbf{M} \rangle_{\mathrm{F}}$:
\begin{align*}
\langle \mathbf{M}, \mathbf{M} \rangle_{\mathrm{F}} =& \langle \mathrm{diag}(\mathbf{d}) + \mathrm{diag}(\mathbf{w}) K(\mathbf{a}) \mathrm{diag}(\mathbf{w}), \mathrm{diag}(\mathbf{d}) + \mathrm{diag}(\mathbf{w}) K(\mathbf{a}) \mathrm{diag}(\mathbf{w}) \rangle_{\mathrm{F}} \\
=& \langle \mathrm{diag}(\mathbf{d}), \mathrm{diag}(\mathbf{d}) \rangle_{\mathrm{F}} + 2 \langle \mathrm{diag}(\mathbf{d}), \mathrm{diag}(\mathbf{w}) K(\mathbf{a}) \mathrm{diag}(\mathbf{w}) \rangle_{\mathrm{F}} \nonumber \\ 
&+ \langle \mathrm{diag}(\mathbf{w}) K(\mathbf{a}) \mathrm{diag}(\mathbf{w}), \mathrm{diag}(\mathbf{w}) K(\mathbf{a}) \mathrm{diag}(\mathbf{w}) \rangle_{\mathrm{F}}.
\end{align*}

Evaluating each component individually:
\begin{align*}
\langle \mathrm{diag}(\mathbf{d}), \mathrm{diag}(\mathbf{d}) \rangle_{\mathrm{F}} &= \|\mathbf{d}\|_2^2, \\
2 \langle \mathrm{diag}(\mathbf{d}), \mathrm{diag}(\mathbf{w}) K(\mathbf{a}) \mathrm{diag}(\mathbf{w}) \rangle_{\mathrm{F}} &= 2 \sum_{j} d_j w_j^2 (K(\mathbf{a}))_{jj} = 2\, \mathbf{d}^\top \mathbf{w}^2, \\
\langle \mathrm{diag}(\mathbf{w}) K(\mathbf{a}) \mathrm{diag}(\mathbf{w}), \mathrm{diag}(\mathbf{w}) K(\mathbf{a}) \mathrm{diag}(\mathbf{w}) \rangle_{\mathrm{F}} &= \sum_{i,j} w_i^2 w_j^2 (K(\mathbf{a})_{ij})^2 = (\mathbf{w}^2)^\top K(2\mathbf{a}) (\mathbf{w}^2).
\end{align*}
The kernel self-inner product relies on $(e^{-|a_i - a_j|})^2 = e^{-2|a_i - a_j|} = K(2\mathbf{a})_{ij}$. Thus, the complete model term is:
\begin{equation*}
\langle \mathbf{M}, \mathbf{M} \rangle_{\mathrm{F}} = \|\mathbf{d}\|_2^2 + 2\, \mathbf{d}^\top \mathbf{w}^2 + (\mathbf{w}^2)^\top K(2\mathbf{a}) (\mathbf{w}^2).
\end{equation*}

Next, we evaluate the data-alignment term using the identity $\langle \mathbf{M}, \mathbf{v}_i \mathbf{v}_i^\top \rangle_{\mathrm{F}} = \mathbf{v}_i^\top \mathbf{M} \mathbf{v}_i$:
\begin{align*}
\mathbf{v}_i^\top \mathbf{M} \mathbf{v}_i =& \mathbf{v}_i^\top \left( \mathrm{diag}(\mathbf{d}) + \mathrm{diag}(\mathbf{w}) K(\mathbf{a}) \mathrm{diag}(\mathbf{w}) \right) \mathbf{v}_i \\
=& \mathbf{v}_i^\top \mathrm{diag}(\mathbf{d}) \mathbf{v}_i + \mathbf{v}_i^\top \mathrm{diag}(\mathbf{w}) K(\mathbf{a}) \mathrm{diag}(\mathbf{w}) \mathbf{v}_i \\
=& \mathbf{d}^\top \mathbf{v}_i^2 + (\mathbf{w} \odot \mathbf{v}_i)^\top K(\mathbf{a}) (\mathbf{w} \odot \mathbf{v}_i).
\end{align*}

Summing over the batch yields the final tractable objective function:
\begin{equation*}
\mathcal{L}_{\mathrm{F}} = \|\mathbf{d}\|_2^2 + 2\, \mathbf{d}^\top \mathbf{w}^2 + (\mathbf{w}^2)^\top K(2\mathbf{a}) (\mathbf{w}^2) - \frac{2}{N} \sum_{i=1}^N \left( \mathbf{d}^\top \mathbf{v}_i^2 + (\mathbf{w} \odot \mathbf{v}_i)^\top K(\mathbf{a}) (\mathbf{w} \odot \mathbf{v}_i) \right).
\end{equation*}

\section{Quasi-Linear Time Evaluation of the Laplace Kernel via Prefix and Suffix Sums}\label{appendix:sec_prefix_suffix}

In the main text, we claim that the quadratic form $\mathbf{x}^\top K(\mathbf{a}) \mathbf{x}$ can be evaluated in $\mathcal{O}(C \log{C})$ time and linear $\mathcal{O}(C)$ memory without ever constructing the dense $C\!\times\! C$ kernel matrix $K(\mathbf{a})$. This efficiency is achieved by exploiting the multiplicative separability of the 1D Laplace kernel to compute the matrix-vector product $\mathbf{y} = K(\mathbf{a})\mathbf{x}$ using recursive prefix and suffix sums.

\paragraph{Formulating the Matrix-Vector Product.}
As established in Appendix~\ref{appendix:sec_gmrf}, we first apply a sorting permutation $\pi$ such that the latent coordinates are monotonically increasing: $a_{\pi(1)} \le a_{\pi(2)} \le \dots \le a_{\pi(C)}$. Let $\tilde{\mathbf{a}}$ and $\tilde{\mathbf{x}}$ denote the permuted coordinate and input vectors, respectively. The $i$-th element of the matrix-vector product $\tilde{\mathbf{y}} = K(\tilde{\mathbf{a}})\tilde{\mathbf{x}}$ is given by the sum:
\begin{equation*}
    \tilde{y}_i = \sum_{j=1}^C e^{-|\tilde{a}_i - \tilde{a}_j|} \tilde{x}_j.
\end{equation*}
By explicitly splitting this sum into the elements occurring before $i$ (the prefix), the element exactly at $i$, and the elements occurring after $i$ (the suffix), we can drop the absolute value:
\begin{equation*}
    \tilde{y}_i = \underbrace{\sum_{j=1}^{i-1} e^{-(\tilde{a}_i - \tilde{a}_j)} \tilde{x}_j}_{\text{Prefix: } L_i} + \tilde{x}_i + \underbrace{\sum_{j=i+1}^C e^{-(\tilde{a}_j - \tilde{a}_i)} \tilde{x}_j}_{\text{Suffix: } R_i}.
\end{equation*}

\paragraph{Recursive Prefix and Suffix Evaluation.}
Because the exponential function is multiplicatively separable (i.e., $e^{-(\tilde{a}_i - \tilde{a}_j)} = e^{-(\tilde{a}_i - \tilde{a}_{i-1})} e^{-(\tilde{a}_{i-1} - \tilde{a}_j)}$), we can compute the prefix sum $L_i$ and suffix sum $R_i$ recursively, entirely avoiding nested loops.

We define the spatial decay factor between adjacent sorted coordinates as $\rho_i = e^{-(\tilde{a}_{i+1} - \tilde{a}_i)}$. The left-to-right prefix sum $\mathbf{L}$ is computed via a single forward pass:
\begin{equation*}
    L_1 = 0, \quad L_i = \rho_{i-1} (L_{i-1} + \tilde{x}_{i-1}) \quad \text{for } i = 2, \dots, C.
\end{equation*}

Similarly, the right-to-left suffix sum $\mathbf{R}$ is computed via a single backward pass:
\begin{equation*}
    R_C = 0, \quad R_i = \rho_{i} (R_{i+1} + \tilde{x}_{i+1}) \quad \text{for } i = C-1, \dots, 1.
\end{equation*}

\paragraph{Final Assembly and Complexity.}
Once the recursive passes are complete, the output vector $\tilde{\mathbf{y}}$ is simply assembled via an element-wise addition:
\begin{equation*}
    \tilde{\mathbf{y}} = \mathbf{L} + \tilde{\mathbf{x}} + \mathbf{R}.
\end{equation*}
We then apply the inverse permutation $\pi^{-1}$ to restore the original channel ordering, yielding $\mathbf{y} = K(\mathbf{a})\mathbf{x}$. Finally, the quadratic form is computed via a standard dot product:
\begin{equation*}
    \mathbf{x}^\top K(\mathbf{a}) \mathbf{x} = \mathbf{x}^\top \mathbf{y}.
\end{equation*}

By replacing the dense matrix–vector multiplication with a forward pass, a backward pass, and an element-wise sum, the computational complexity is reduced from $\mathcal{O}(C^2)$ to $\mathcal{O}(C \log C)$, where the $\log C$ term arises from sorting. Furthermore, because the memory footprint only requires storing the 1D vectors $\mathbf{L}$, $\mathbf{R}$, and $\boldsymbol{\rho}$, the memory complexity is bounded to $\mathcal{O}(C)$.

\section{Gaussian Markov Random Field and the Tridiagonal Precision Matrix}
\label{appendix:sec_gmrf}

In this section, we provide a complete derivation showing that the LCM induces a sparse, tridiagonal precision matrix. This result follows from the equivalence between the Laplace kernel on a 1D coordinate system and a Gaussian Markov Random Field (GMRF).

\paragraph{1D Ordering and Markov Structure.}
The kernel component of our model is defined as $(K(\mathbf{a}))_{ij} = e^{-|a_i - a_j|}$, where $\mathbf{a}$ denotes learned 1D coordinates. Without loss of generality, we assume the coordinates are sorted such that $a_1 \le a_2 \le \dots \le a_C$.

Define $\rho_i = e^{-(a_{i+1} - a_i)}$ as the correlation between adjacent channels. For any $i < j$, the distance decomposes as
\begin{equation*}
    |a_i - a_j| = \sum_{k=i}^{j-1} (a_{k+1} - a_k),
\end{equation*}
which implies a multiplicative factorization of the kernel:
\begin{equation*}
    \mathbf{K}_{ij} = e^{-\sum_{k=i}^{j-1} (a_{k+1} - a_k)} = \prod_{k=i}^{j-1} \rho_k, \quad \text{for } i \le j.
\end{equation*}
By symmetry, $\mathbf{K}_{ji} = \mathbf{K}_{ij}$. This decomposition reveals a Markov structure: correlations between distant channels are mediated through intermediate ones.

\paragraph{Precision Matrix via Autoregressive Representation.}
Let $\mathbf{x} \sim \mathcal{N}(\mathbf{0}, \mathbf{K})$, where $\mathbf{K}$ is a correlation matrix induced by the Laplace kernel. By construction, its diagonal entries satisfy $K_{i,i} = 1$, implying unit marginal variance for all variables, i.e., $\mathrm{Var}(x_i) = 1$.

Due to the Markov structure of the 1D exponential kernel, $\mathbf{x}$ admits an equivalent autoregressive (AR(1)) representation:
\begin{align*}
    x_1 &= \epsilon_1, \quad &\epsilon_1 \sim \mathcal{N}(0, 1), \\
    x_i &= \rho_{i-1} x_{i-1} + \epsilon_i, \quad &\epsilon_i \sim \mathcal{N}(0, 1 - \rho_{i-1}^2), \quad i \ge 2,
\end{align*}
where $\rho_{i-1}$ encodes the correlation between consecutive variables.

The variance of the innovations $\epsilon_i$ follows directly from the unit marginal constraint. Applying the variance operator to $x_i = \rho_{i-1} x_{i-1} + \epsilon_i$ and using independence yields
\begin{equation*}
    \mathrm{Var}(x_i) = \rho_{i-1}^2 \mathrm{Var}(x_{i-1}) + \mathrm{Var}(\epsilon_i).
\end{equation*}
Substituting $\mathrm{Var}(x_i) = \mathrm{Var}(x_{i-1}) = 1$ gives $\mathrm{Var}(\epsilon_i) = 1 - \rho_{i-1}^2$.

This system can be written compactly as $\mathbf{L}\mathbf{x} = \boldsymbol{\epsilon}$, where $\mathbf{L}$ is a lower bidiagonal matrix:
\begin{equation*}
\mathbf{L} =
\begin{bmatrix}
    1 & 0 & 0 & \dots & 0 \\
    -\rho_1 & 1 & 0 & \dots & 0 \\
    0 & -\rho_2 & 1 & \dots & 0 \\
    \vdots & \ddots & \ddots & \ddots & \vdots \\
    0 & \dots & 0 & -\rho_{C-1} & 1
\end{bmatrix},
\end{equation*}
and $\boldsymbol{\epsilon}$ has diagonal covariance
\begin{equation*}
\mathbf{D} = \mathrm{diag}(1, \, 1-\rho_1^2, \, \dots, \, 1-\rho_{C-1}^2).
\end{equation*}

It follows that the covariance and precision matrices admit the factorization
\begin{equation*}
\mathbf{K} = \mathbf{L}^{-1} \mathbf{D} \mathbf{L}^{-\top}, 
\quad
\mathbf{Q} = \mathbf{K}^{-1} = \mathbf{L}^\top \mathbf{D}^{-1} \mathbf{L}.
\end{equation*}

\paragraph{Explicit Form of the Precision Matrix.}
The inverse covariance $\mathbf{Q}$ is given by $\mathbf{Q} = \mathbf{L}^\top \mathbf{D}^{-1} \mathbf{L}$, where
\begin{equation*}
    \mathbf{D}^{-1}_{1,1} = 1, \quad \mathbf{D}^{-1}_{k,k} = \frac{1}{1 - \rho_{k-1}^2}, \quad 1 < k \le C.
\end{equation*}

For diagonal entries:
\begin{itemize}
    \item $i = 1$:
    \begin{equation*}
        \mathbf{Q}_{1,1} = \frac{1}{1 - \rho_1^2}.
    \end{equation*}
    \item $1 < i < C$:
    \begin{equation*}
        \mathbf{Q}_{i,i} =
        \frac{1}{1 - \rho_{i-1}^2} + \frac{\rho_i^2}{1 - \rho_i^2}.
    \end{equation*}
    \item $i = C$:
    \begin{equation*}
        \mathbf{Q}_{C,C} = \frac{1}{1 - \rho_{C-1}^2}.
    \end{equation*}
\end{itemize}

For off-diagonal entries, by symmetry it suffices to consider $i < j$:
\begin{equation*}
    \mathbf{Q}_{i,j} = \sum_{k=1}^C \mathbf{L}_{k,i} \mathbf{D}^{-1}_{k,k} \mathbf{L}_{k,j}.
\end{equation*}
Due to the bidiagonal structure:
\begin{itemize}
    \item If $j = i+1$:
    \begin{equation*}
        \mathbf{Q}_{i,i+1} = -\frac{\rho_i}{1 - \rho_i^2}.
    \end{equation*}
    \item If $|i - j| > 1$:
    \begin{equation*}
        \mathbf{Q}_{i,j} = 0.
    \end{equation*}
\end{itemize}

This derivation shows that the precision matrix $\mathbf{Q}$ is strictly tridiagonal. Consequently, the Laplace kernel over 1D coordinates induces a GMRF with local dependencies, enabling efficient computation while preserving structured correlations.

\section{Generator Training Details}
\label{appendix:sec_generator_training_details}

In DFCIL, the generator acts as a synthesizer that produces pseudo-samples $\mathbf{X}_{\text{old}}$ for previously learned tasks. Following the PMI framework, we do not train an auxiliary parametric generative model (e.g., a GAN), but instead instantiate the generator through a \emph{per-layer model inversion} procedure. In this formulation, intermediate feature tensors are treated as optimization variables, enabling the reconstruction of high-fidelity inputs by propagating constraints backward through the frozen network in a layer-wise manner. 

Building on this paradigm, we extend PMI with \our{}, enabling explicit modeling of structured correlations within feature representations during inversion and overcoming the diagonal covariance assumptions used in prior work.

\paragraph{Contrastive Feature Selection (CFS).}
A key challenge in model inversion is defining informative and diverse semantic targets. In standard classification models, mapping a high-dimensional feature vector to a scalar label discards a substantial amount of information. As a result, relying solely on class centroids as inversion targets often leads to mode collapse and synthetic samples that lack diversity.

To preserve the geometry of the original feature distribution, we follow PMI and adopt CFS. For each previously learned class $c$, we approximate the feature distribution at the penultimate layer using a Gaussian model $\mathcal{N}(\boldsymbol{\mu}_c, \boldsymbol{\sigma}_c^2)$. 

We then train a lightweight contrastive MLP $f_{\mathrm{cont}}$ on real features using a negative contrastive objective:
\begin{equation*}
    \mathcal{L}_{\mathrm{cont}}(\mathbf{o}_{L,i}, \mathcal{S}_{\mathrm{neg}}; f_{\mathrm{cont}}) =
    \log \mathbb{E}_{\mathbf{o}_{L,j} \in \mathcal{S}_{\mathrm{neg}}}
    \left[
    e^{\cos(f_{\mathrm{cont}}(\mathbf{o}_{L,i}), f_{\mathrm{cont}}(\mathbf{o}_{L,j}))}
    \right],
\end{equation*}
where $\mathcal{S}_{\mathrm{neg}}$ denotes a negative set and $\cos(\cdot,\cdot)$ is cosine similarity. Optimizing this objective encourages features to be uniformly distributed on a hypersphere, ensuring that no semantic subspace is overly compressed.

During generation, we sample a large candidate pool from $\mathcal{N}(\boldsymbol{\mu}_c, \boldsymbol{\sigma}_c^2)$ and apply a greedy selection procedure (see Algorithm~2 in \cite{tongmodel}). At each step, we select features that minimize the contrastive loss with respect to already chosen samples. This produces a subset $\mathcal{S}_{\mathrm{feat}}$ that maximizes diversity and distributional coverage, providing robust semantic anchors for inversion.

\paragraph{Linking Generation to Per-layer Inversion.}
We decompose the feature extractor $f_{\mathrm{feat}}$ into a sequence of $L$ blocks (e.g., residual blocks for CNNs or transformer blocks for ViTs). Directly optimizing pixels to match high-level semantic representations is highly non-convex. To address this, we decompose the problem into a sequence of tractable layer-wise sub-problems.

Let $\mathbf{o}_l$ denote the input to block $l$, and $\mathbf{o}_{l+1}$ its output. The inversion proceeds backward from the deepest layer $L$ to the input:
\begin{enumerate}
    \item Obtain target features $\mathcal{S}_{\mathrm{feat}}$ at layer $L$ using CFS.
    \item Initialize $\hat{\mathbf{o}}_l$ as Gaussian noise scaled by stored input statistics.
    \item Optimize $\hat{\mathbf{o}}_l$ via gradient descent so that its forward pass through the frozen block matches the target.
    \item Use the optimized $\hat{\mathbf{o}}_l$ as the target for the preceding layer $l-1$.
\end{enumerate}
This sequential procedure decomposes a highly non-convex global objective into a series of well-conditioned local problems.

\paragraph{Layer-wise Optimization Objective.}
At each block $l > 0$, we optimize $\hat{\mathbf{o}}_l$ using
\begin{equation*}
    \mathcal{L}_{\mathrm{layer}}^{(l)} =
    \alpha_{\mathrm{match}}\mathcal{L}^{(l)}_{\mathrm{match}} +
    \alpha_{\mathrm{stat}}\mathcal{L}^{(l)}_{\mathrm{stat}} +
    \alpha_{\mathrm{in}}\mathcal{L}^{(l)}_{\mathrm{in}}.
\end{equation*}

\textbf{1. Feature Matching Loss ($\mathcal{L}^{(l)}_{\mathrm{match}}$).}
The formulation of the deterministic matching loss depends strictly on the depth of the layer being optimized. 

At the topmost layer ($l=L$), the optimization aims to align the generated features directly with the target class label $y$. Therefore, the matching objective utilizes a standard CE loss: $\mathcal{L}^{(L)}_{\mathrm{match}} = \mathcal{L}_{\mathrm{ce}}(\hat{\mathbf{o}}_L, y)$.

However, for all intermediate layers ($l < L$), the optimization relies on the cascading targets from the layer above. Here, the objective transitions to a Mean Squared Error (MSE) loss, which enforces consistency with the localized target features propagated backward from the previously optimized layer:
\begin{equation*}
    \mathcal{L}^{(l)}_{\mathrm{match}} =
    \big\|
    \mathrm{norm}(\mathrm{Block}_l(\hat{\mathbf{o}}_l))
    -
    \mathcal{S}_{\mathrm{feat}}^{(l+1)}
    \big\|_2^2.
\end{equation*}
Specifically, $\hat{\mathbf{o}}_l$ is the intermediate feature representation being optimized, which serves as the input to $\mathrm{Block}_l$. $\mathrm{Block}_l$ refers to a primary structural unit of the frozen network, $\mathrm{norm}(\cdot)$ denotes the normalization applied to the block's output, and $\mathcal{S}_{\mathrm{feat}}^{(l+1)}$ represents the fixed target features obtained from the previously optimized layer $l+1$.

\textbf{2. Distribution Statistic Loss ($\mathcal{L}^{(l)}_{\mathrm{stat}}$):}

We regularize intermediate activations to match the feature statistics observed on real data by modeling their distribution with the proposed LCM and minimizing the exact Gaussian Negative Log-Likelihood (NLL).

Given a batch of generated features $\{\hat{\mathbf{o}}_{l,i}\}_{i=1}^N \in \mathbb{R}^C$, the objective under the multivariate Gaussian distribution $\mathcal{N}(\boldsymbol{\mu}^{(l)}, \boldsymbol{\Sigma}^{(l)})$ is:
\begin{equation}
    \mathcal{L}^{(l)}_{\mathrm{stat}} = \frac{1}{2N} \sum_{i=1}^N \left( \log \det \boldsymbol{\Sigma}^{(l)} + (\hat{\mathbf{o}}_{l,i} - \boldsymbol{\mu}^{(l)})^\top (\boldsymbol{\Sigma}^{(l)})^{-1} (\hat{\mathbf{o}}_{l,i} - \boldsymbol{\mu}^{(l)}) \right).
    \label{eq:dense_nll}
\end{equation}

Direct evaluation requires $\mathcal{O}(C^3)$ time, which is prohibitive for high-dimensional features. To bypass this, we exploit the structure of the LCM covariance.

\textbf{Generative Matrix Factorization.}
A random vector distributed as $\mathcal{N}(\boldsymbol{\mu}^{(l)}, \boldsymbol{\Sigma}^{(l)})$ can be constructively generated by drawing a latent vector $\mathbf{z} \sim \mathcal{N}(\mathbf{0}, \mathbf{K})$ and an independent noise vector $\boldsymbol{\eta} \sim \mathcal{N}(\mathbf{0}, \mathbf{D})$, such that:
\begin{equation}
    \hat{\mathbf{o}}_{l,i} - \boldsymbol{\mu}^{(l)} = \mathbf{W}\mathbf{z} + \boldsymbol{\eta}_{i}.
    \label{eq:generative_matrix}
\end{equation}
Eq.~\ref{eq:generative_matrix} yields the exact covariance $\mathbb{E}[(\mathbf{W}\mathbf{z} + \boldsymbol{\eta})(\mathbf{W}\mathbf{z} + \boldsymbol{\eta})^\top] = \mathbf{W}\mathbf{K}\mathbf{W}^\top + \mathbf{D} = \boldsymbol{\Sigma}^{(l)}$. This factorization reveals that the final generated features are simply a standard correlated latent process ($\mathbf{z}$), scaled element-wise by the diagonal weights ($\mathbf{W}$), with added noise ($\boldsymbol{\eta}$).

\textbf{Mapping to a 1D State-Space Model.}
Recall from Eq.~\ref{eq:generative_matrix} that the generated features are driven by a latent random vector $\mathbf{z} \sim \mathcal{N}(\mathbf{0}, \mathbf{K})$. Conceptually, $\mathbf{z}$ represents an unobserved, normalized hidden signal that correlates all the feature channels together. 

Because the correlation matrix $\mathbf{K}$ relies purely on the $L_1$ distance between the learned channel coordinates $\mathbf{a}$, this underlying continuous signal strictly satisfies the Markov property—provided we evaluate the channels in the correct spatial order. Therefore, we apply a permutation $\pi$ that sorts the channels such that their coordinates are strictly increasing: $a_{\pi(1)} \le \dots \le a_{\pi(C)}$. 

Under this sorted ordering, we can break down the dense vector equation (Eq.~\ref{eq:generative_matrix}) into a step-by-step scalar sequence. Let $z_k$ denote the $k$-th scalar element of the sorted hidden signal. Similarly, let $x_{i,k} = \hat{o}_{l,i,\pi(k)}$ be the actual observed feature for the $k$-th sorted channel, with corresponding historical parameters $\mu_k = \mu_{\pi(k)}^{(l)}$, $w_k = w_{\pi(k)}^{(l)}$, and $u_k = u_{\pi(k)}^{(l)}$. 

The spatial distance between adjacent channels is $\Delta a_k = a_{\pi(k)} - a_{\pi(k-1)}$. This distance dictates how strongly the hidden signal decays from one channel to the next, defining the autoregressive transition $\phi_k = \exp(-\Delta a_k)$ and the injected process noise $q_k = 1 - \phi_k^2$. The dense multivariate model is thus mathematically equivalent to a 1D linear state-space model (SSM):
\begin{align}
    z_k &= \phi_k z_{k-1} + \epsilon_k, \quad \epsilon_k \sim \mathcal{N}(0, q_k), \label{eq:ssm_hidden} \\
    x_{i,k} &= \mu_k + w_k z_k + \eta_{i,k}, \quad \eta_{i,k} \sim \mathcal{N}(0, u_k). \label{eq:ssm_obs}
\end{align}

Eq.~\ref{eq:ssm_hidden} describes how the unobserved hidden signal $z_k$ evolves smoothly across the sorted channels. Eq.~\ref{eq:ssm_obs} demonstrates how the actual observed feature $x_{i,k}$ is generated: by scaling the hidden signal by the weight $w_k$, shifting it by the historical mean $\mu_k$, and adding independent channel noise $\eta_{i,k}$.

\textbf{Exact Likelihood via Kalman Filtering.}
To compute the exact likelihood of the entire sequence, we use a Kalman filter. The filter moves step-by-step through the sorted channels, maintaining a running estimate of the true hidden signal. At the beginning of step $k$, our knowledge of the previous hidden state $z_{k-1}$ is represented by a Gaussian distribution with mean $m_{i,k-1}$ and variance $P_{k-1}$. The filtering process at each channel follows three distinct stages.

Before observing the $k$-th channel, we use the autoregressive transition (Eq.~\ref{eq:ssm_hidden}) to predict the current hidden state. The expected mean and variance of this prior prediction are:
\begin{align*}
    m_{i,k|k-1} &= \phi_k m_{i,k-1}, \\
    P_{k|k-1} &= \phi_k^2 P_{k-1} + q_k. 
\end{align*}

Next, we predict what the actual observed feature $x_{i,k}$ should be, based on our hidden state prediction. The difference between the true observation and our expectation is called the \textit{innovation} (or prediction error) $v_{i,k}$:
\begin{equation*}
    v_{i,k} = (x_{i,k} - \mu_k) - w_k \underbrace{m_{i,k|k-1}}_{\text{Predicted State}}.
\end{equation*}
Because both the hidden state and the channel observation noise are uncertain, the variance of this expected observation, denoted as $S_k$, is the sum of both uncertainties:
\begin{equation*}
    S_k = w_k^2 \underbrace{P_{k|k-1}}_{\text{State Variance}} + u_k.
\end{equation*}

Finally, to prepare for the next step $k+1$, the filter corrects its internal estimate of the hidden state using the newly observed prediction error $v_{i,k}$. The degree of this correction is controlled by the Kalman gain $K_k = P_{k|k-1} w_k / S_k$, which acts as a dynamic ``trust ratio.'' It balances how much the filter trusts its own prior prediction ($P_{k|k-1}$) versus the noisy new observation ($S_k$). 

Using this gain, we compute the updated posterior mean and variance:
\begin{align}
    m_{i,k} &= m_{i, k|k-1} + K_k v_{i,k}, \label{eq:kf_update_mean} \\
    P_k &= (1 - K_k w_k) P_{k|k-1}. \label{eq:kf_update_var}
\end{align}
Eq.~\ref{eq:kf_update_mean} intuitively shows that the new mean is simply the old prediction shifted by a fraction of the error. Eq.~\ref{eq:kf_update_var} demonstrates that observing new data always reduces our uncertainty: the prior variance $P_{k|k-1}$ is strictly shrunk by the factor $(1 - K_k w_k)$.

The primary advantage of this sequential formulation is how elegantly it simplifies the likelihood calculation. Direct evaluation of the full-covariance multivariate Gaussian requires calculating a massive, tangled joint probability. However, the Kalman filter mathematically untangles this vector into a sequence of purely independent ``surprises''—the innovations $v_{i,k}$.

Because each innovation $v_{i,k}$ captures only the orthogonal, completely new information arriving at channel $k$, these errors are statistically independent from one another and follow a simple 1D Gaussian distribution: $v_{i,k} \sim \mathcal{N}(0, S_k)$.

By the rules of probability, the joint likelihood of independent events is simply the product of their individual likelihoods. Consequently, the computationally heavy $\mathcal{O}(C^3)$ dense matrix objective from Eq.~\ref{eq:dense_nll} algebraically collapses into a sum of standard 1D NLLs. This allows us to evaluate the exact loss in $\mathcal{O}(C\log{C})$ time:
\begin{equation*}
    \mathcal{L}^{(l)}_{\mathrm{stat}} = \frac{1}{2N} \sum_{i=1}^N \sum_{k=1}^C \left( \log(2\pi S_k) + \frac{v_{i,k}^2}{S_k} \right).
\end{equation*}

\textbf{3. Input Statistic Prior ($\mathcal{L}^{(l)}_{\mathrm{in}}$):}
Due to the strong non-linearity of deep networks, directly optimizing the input tensor $\hat{\mathbf{o}}_l$ to match downstream targets can lead to adversarial or out-of-distribution activations. Although such features may satisfy the feature-matching objective, they often lack meaningful structure and can destabilize subsequent inversion steps.

To mitigate this issue, we constrain $\hat{\mathbf{o}}_l$ using a Gaussian prior defined over the input activations of block $l$. During training on tasks $1{:}t$, we record the empirical channel-wise mean and variance of these activations, denoted as $\boldsymbol{\mu}_{1:t}^{(l,\mathrm{in})}$ and $(\boldsymbol{\sigma}_{1:t}^{(l,\mathrm{in})})^2$. 

During inversion, we enforce the empirical batch statistics of the optimized tensor ($\hat{\boldsymbol{\mu}}^{(l)}$ and $(\hat{\boldsymbol{\sigma}}^{(l)})^2$) to match these stored values. This is achieved by minimizing the KL divergence between the current batch distribution and the historical Gaussian prior, applied independently across channels:
\begin{equation*}
    \mathcal{L}^{(l)}_{\mathrm{in}} =
    \frac{1}{2C}
    \sum_{c=1}^C
    \left(
    \log \frac{(\sigma_{1:t,c}^{(l,\mathrm{in})})^2}{(\hat{\sigma}_{c}^{(l)})^2}
    +
    \frac{(\hat{\sigma}_{c}^{(l)})^2 + (\hat{\mu}_{c}^{(l)} - \mu_{1:t,c}^{(l,\mathrm{in})})^2}{(\sigma_{1:t,c}^{(l,\mathrm{in})})^2}
    - 1
    \right).
\end{equation*}

This channel-wise prior anchors the optimization to the distribution of features naturally observed at the input of block $l$, preventing drift toward unrealistic activations and stabilizing the layer-wise inversion process.

\paragraph{Pixel-Level Optimization.}
The final output of the generator must be replay images $\hat{\mathbf{x}} \in \mathbb{R}^{3 \times H \times W}$; therefore, the last stage of layer-wise inversion ($l=0$) operates directly in pixel space. However, optimizing pixels solely to match high-level semantic features often yields high-frequency, adversarial artifacts that are unsuitable for continual learning. To mitigate this, we augment the objective with an image prior $\mathcal{L}_{\mathrm{pr}}$ regularization, which enforces spatial smoothness by penalizing local differences:
\begin{equation*}
    \mathcal{L}_{\mathrm{pr}}(\hat{\mathbf{x}}) =
    \sum_{c=1}^3 \sum_{h=1}^{H-1} \sum_{w=1}^{W-1}
    \left(
    (\hat{x}_{c,h+1,w} - \hat{x}_{c,h,w})^2 +
    (\hat{x}_{c,h,w+1} - \hat{x}_{c,h,w})^2
    \right).
\end{equation*}

\paragraph{Full-Model Refinement and Pseudo-Sample Generation.}
Although layer-wise inversion decomposes a highly non-convex optimization problem into tractable sub-problems, small approximation errors inevitably accumulate across layers. To correct these errors, the intermediate reconstructions $\hat{\mathbf{x}}$ are not treated as final outputs, but instead serve as strong initializations for a subsequent end-to-end refinement stage.

During this phase, the reconstructed inputs $\hat{\mathbf{x}}$ are optimized directly through the entire frozen feature extractor $f_{\mathrm{feat}}$. Unlike the sequential inversion stage, intermediate layer-specific targets ($\mathcal{S}_{\mathrm{feat}}^{(l+1)}$) are no longer available, and therefore the layer-wise reconstruction losses $\mathcal{L}^{(l)}_{\mathrm{mse}}$ are discarded.

The refinement objective instead combines the final-layer CE loss $\mathcal{L}_{\mathrm{ce}}$, which preserves target class semantics, with the proposed full-covariance Distribution Statistic Loss $\mathcal{L}^{(l)}_{\mathrm{stat}}$ evaluated across intermediate representations. In this setting, the statistical prior replaces explicit layer-wise supervision and constrains the generated samples to remain consistent with the learned feature distribution. This final global optimization stage compensates for accumulated approximation errors and produces pseudo-samples $\mathbf{X}_{\text{old}}$ with improved structural and semantic coherence for replay.

\section{Feature Extractor and Classifier Training Details}
\label{appendix:sec_feature_extractor_and_classifier_training_details}

In this section, we provide a detailed description of the feature extractor and classifier training protocol used in our method. The procedure follows the two-stage optimization strategy introduced in R-DFCIL and adopted in our \our{} framework. We emphasize both the formal definitions and the underlying intuition behind each component.

\paragraph{Problem Formulation and Notation.}
Let $f_{\mathrm{feat}}(\cdot; \boldsymbol{\theta})$ denote the feature extractor and $f_{\mathrm{head}}(\cdot; \boldsymbol{\phi})$ denote the linear classification head. Assume the model has already been trained on tasks $1$ through $t$. We retain a frozen copy of this historical model, parameterized by $(\boldsymbol{\theta}_{1:t}, \boldsymbol{\phi}_{1:t})$, which serves as a teacher during subsequent training.

When learning a new task $t+1$, the model updates its active parameters $(\boldsymbol{\theta}, \boldsymbol{\phi})$. To accommodate the new label space $\mathcal{Y}_{t+1}$, the classifier is expanded with additional parameters, denoted by $\boldsymbol{\phi}_{\text{new}}$, while the parameters corresponding to previously learned classes are denoted by $\boldsymbol{\phi}_{\text{old}}$.

A key challenge in this setting is catastrophic forgetting caused by the absence of past data. To mitigate this, the training process interleaves two types of data: (i) real samples from the current task, $(\mathbf{X}_{\text{new}}, Y_{\text{new}}) \subset \mathcal{D}_{t+1}$, and (ii) synthesized pseudo-samples from previous tasks, $(\mathbf{X}_{\text{old}}, Y_{\text{old}})$.

\paragraph{Stage 1: Relation-Guided Representation Learning (RRL).}
In the first stage, we jointly optimize the feature extractor and classifier parameters $(\boldsymbol{\theta}, \boldsymbol{\phi})$ using the relation-guided representation learning objective:
\begin{equation*}
    \mathcal{L}_{\mathrm{rrl}} =
    \lambda_{\mathrm{lce}} \mathcal{L}_{\mathrm{lce}} +
    \lambda_{\mathrm{hkd}} \mathcal{L}_{\mathrm{hkd}} +
    \lambda_{\mathrm{rkd}} \mathcal{L}_{\mathrm{rkd}}.
\end{equation*}

These three components are designed to balance the competing objectives of learning new information (plasticity) and preserving previously acquired knowledge (stability).

\textbf{1. Local Classification Loss ($\mathcal{L}_{\mathrm{lce}}$).}
The local classification loss is a standard CE objective applied exclusively to real samples from the current task. By restricting supervision to $\mathbf{X}_{\text{new}}$, the model learns new classes without being biased by imperfections in synthetic data:
\begin{equation*}
    \mathcal{L}_{\mathrm{lce}} =
    \frac{1}{|\mathbf{X}_{\text{new}}|}
    \sum_{(\mathbf{x}, y) \in (\mathbf{X}_{\text{new}}, Y_{\text{new}})}
    \mathcal{L}_{\mathrm{ce}}\big(
    \mathrm{softmax}(f_{\mathrm{head}}(f_{\mathrm{feat}}(\mathbf{x}; \boldsymbol{\theta}); \boldsymbol{\phi}_{\text{new}})), y
    \big).
\end{equation*}

\textbf{2. Hard Knowledge Distillation ($\mathcal{L}_{\mathrm{hkd}}$).}
To explicitly preserve knowledge from previous tasks, we apply hard knowledge distillation on synthetic samples. This term enforces consistency between the outputs of the current model and the frozen teacher by penalizing deviations in logits:
\begin{equation*}
    \mathcal{L}_{\mathrm{hkd}} =
    \frac{1}{|\mathbf{X}_{\text{old}}| \, |\mathcal{Y}_{1:t}|}
    \sum_{\mathbf{x} \in \mathbf{X}_{\text{old}}}
    \left\|
    f_{\mathrm{head}}(f_{\mathrm{feat}}(\mathbf{x}; \boldsymbol{\theta}_{1:t}); \boldsymbol{\phi}_{1:t})
    -
    f_{\mathrm{head}}(f_{\mathrm{feat}}(\mathbf{x}; \boldsymbol{\theta}); \boldsymbol{\phi}_{\text{old}})
    \right\|_1.
\end{equation*}

\textbf{3. Relational Knowledge Distillation ($\mathcal{L}_{\mathrm{rkd}}$).}
While $\mathcal{L}_{\mathrm{hkd}}$ constrains absolute predictions, it can overly restrict the feature space. To counterbalance this effect, we introduce relational knowledge distillation, which preserves the geometric structure of the feature space by matching angular relationships between features.

Let $u$ and $v$ be learnable projection functions, and define
$\mathbf{t}_k = u(f_{\mathrm{feat}}(\mathbf{x}_k; \boldsymbol{\theta}_{1:t}))$ and
$\mathbf{s}_k = v(f_{\mathrm{feat}}(\mathbf{x}_k; \boldsymbol{\theta}))$.
For a triplet $(\mathbf{x}_a, \mathbf{x}_b, \mathbf{x}_c)$, the loss is:
\begin{equation*}
    \mathcal{L}_{\mathrm{rkd}} =
    \frac{1}{|\mathbf{X}_{\text{new}}|^3}
    \sum_{\mathbf{x}_a, \mathbf{x}_b, \mathbf{x}_c \in \mathbf{X}_{\text{new}}}
    \left\|
    \cos \angle (\mathbf{t}_a, \mathbf{t}_b, \mathbf{t}_c)
    -
    \cos \angle (\mathbf{s}_a, \mathbf{s}_b, \mathbf{s}_c)
    \right\|_1.
\end{equation*}

\textbf{Adaptive Scaling Factors.}
The contributions of individual loss terms are dynamically adjusted based on the ratio of new to previously learned classes:
\begin{equation*}
    \lambda_{\mathrm{lce}} = \frac{1 + 1/\alpha}{\beta} \lambda_{\mathrm{lce}}^{\text{base}}, \quad
    \lambda_{\mathrm{hkd}} = \alpha \beta \lambda_{\mathrm{hkd}}^{\text{base}}, \quad
    \lambda_{\mathrm{rkd}} = \alpha \beta \lambda_{\mathrm{rkd}}^{\text{base}},
\end{equation*}
where
\begin{equation*}
    \alpha = \log_2\!\left(\frac{|\mathcal{Y}_{t+1}|}{2} + 1\right), \quad
    \beta = \sqrt{\frac{|\mathcal{Y}_{1:t}|}{|\mathcal{Y}_{t+1}|}}.
\end{equation*}

\paragraph{Stage 2: Classification Head Refinement.}
In the second stage, the feature extractor is frozen ($\boldsymbol{\theta}^*$), and only the classifier $\boldsymbol{\phi}$ is updated. The goal of this stage is to correct biases introduced during the first stage and to ensure balanced decision boundaries across all classes.

We optimize a global, class-balanced CE loss over the combined dataset $\mathbf{X}_{\text{all}} = \mathbf{X}_{\text{new}} \cup \mathbf{X}_{\text{old}}$:
\begin{equation*}
    \mathcal{L}_{\mathrm{gce}} =
    \frac{1}{|\mathbf{X}_{\text{all}}|}
    \sum_{(\mathbf{x}, y) \in (\mathbf{X}_{\text{all}}, Y_{\text{all}})}
    \frac{w_y}{\sum_j w_j}
    \mathcal{L}_{\mathrm{ce}}\big(
    \mathrm{softmax}(f_{\mathrm{head}}(f_{\mathrm{feat}}(\mathbf{x}; \boldsymbol{\theta}^*); \boldsymbol{\phi})), y
    \big),
\end{equation*}
where $w_y$ denotes the inverse frequency of class $y$ in the current training data.

\section{Metrics Definition}\label{appendix:sec_metrics_definition}

We evaluate the performance of \our{} using two standard metrics in continual learning: \textit{average accuracy} and \textit{last task accuracy}. Let $a_{t,i}$ denote the classification accuracy on the test set of task $i$ after the model has been trained up to task $t$ (where $i \le t$).

\paragraph{Average Accuracy.}
Average accuracy measures the overall performance of the model across all tasks observed up to a given incremental step $t$. It is defined as
\begin{equation*}
    \mathcal{A}_t = \frac{1}{t} \sum_{i=1}^{t} a_{t,i}.
\end{equation*}
This metric captures the balance between retaining previously learned knowledge and adapting to new tasks. When reporting overall performance, it is common to compute the \textit{average incremental accuracy}, defined as the mean of $\mathcal{A}_t$ over all $T$ incremental steps.

\paragraph{Last Task Accuracy.}
Last task accuracy evaluates the model after completing the full sequence of $T$ tasks. It reflects the final performance across all learned classes and is defined as
\begin{equation*}
    \mathcal{A}_{\mathrm{last}} = \mathcal{A}_T = \frac{1}{T} \sum_{i=1}^{T} a_{T,i}.
\end{equation*}

\section{Implementation Details}\label{appendix:sec_implementation_details}

\paragraph{Benchmarks and Task Splits.}
We strictly follow the experimental protocol of PMI, adopting identical task divisions, class orderings, and evaluation settings to ensure a fair comparison. Each dataset is partitioned into a sequence of disjoint incremental tasks, where new classes are introduced over time without access to data from previous tasks.

For CIFAR-100, we use splits into 5, 10, and 20 tasks in the ResNet-32 setting, and a 10-task split (10 classes per task) in the CLIP-based setting. Tiny-ImageNet is partitioned analogously into 5, 10, and 20 tasks. For CUB-200, we adopt a 10-task split with 20 classes per task, following PMI.

All experiments are conducted in the DFCIL setting, where no real samples from previous tasks are retained. Instead, pseudo-samples are generated via model inversion using stored feature statistics. We follow the exact task configurations of PMI, ensuring a fair and direct comparison.

\paragraph{Grid Search and Optimal Hyperparameters.}
For experiments with the ResNet-32 backbone, we perform a grid search over the regularization weight $\lambda_{\mathrm{F}}$, which controls the contribution of the Frobenius term $\mathcal{L}_{\mathrm{F}}$ (Eq.~\eqref{eq:frobenius_loss_function}), using the range $\{0.01, 0.05, 0.1, 0.25, 0.5, 1.0\}$. For CLIP-based experiments, we adopt a finer grid: $\{0.0001, 0.0005, 0.001, 0.002, 0.005, 0.01, 0.05\}$.

The LCM parameters are optimized using Adam with a learning rate of $0.01$ for $200$ epochs. Despite this number of epochs, optimization remains computationally efficient, as discussed in Appendix~\ref{appendix:sec_additional_experiments}, where we report training time in comparison to PMI.

We adopt the same task splits as defined in the PMI and R-DFCIL protocols. Table~\ref{tab:frob_all} reports the selected optimal values of $\lambda_{\mathrm{F}}$ for each dataset and number of tasks.

\begin{table}[h]
\centering
\caption{Selected $\lambda_{\mathrm{F}}$ values for all experiments.}
\label{tab:frob_all}
\begin{tabular}{l l c c c}
\hline
\textbf{Architecture} & \textbf{Dataset} & \textbf{5 tasks} & \textbf{10 tasks} & \textbf{20 tasks} \\
\hline
ResNet-32 & CIFAR-100     & $0.05$ & $0.5$ & $0.5$  \\
ResNet-32 & Tiny-ImageNet  & $0.1$ & $0.25$ & $0.05$ \\
\hline
CLIP      & CIFAR-100     & -- & $0.0001$ & -- \\
CLIP      & CUB-200       & -- & $0.01$ & -- \\
\hline
\end{tabular}
\end{table}

\paragraph{Code and Reproducibility.}
To ensure a fair comparison, we adopt the same software environment as the PMI codebase. The experiments are implemented in Python using \texttt{PyTorch} (v1.13.1) with \texttt{torchvision} (v0.14.1) and \texttt{timm} (v0.9.7) for model architectures. We rely on \texttt{NumPy} (v1.26.4), \texttt{SciPy} (v1.13.1), and \texttt{scikit-learn} (v1.6.1) for numerical computations, and \texttt{matplotlib} (v3.8.2) and \texttt{tensorboard} (v2.12.1) for visualization and logging. For CLIP-based experiments, we use \texttt{open-clip-torch} (v2.22.0).

All experiments were conducted on NVIDIA A100 (40GB) and RTX 4090 (24GB) GPUs. Full training pipelines, configuration files, and scripts for reproducing all reported results will be made publicly available in our GitHub repository.

\section{Additional Experimental Results}\label{appendix:sec_additional_experiments}

\subsection{ResNet-32 Backbone}

\paragraph{Table~\ref{tab:resnet_cl} with Standard Deviations.}
Table~\ref{appendix:tab_full_resnet_cl} reports average incremental accuracy together with standard deviations (over five runs), corresponding to Table~\ref{tab:resnet_cl} in the main text.

\begin{table}[htbp!]
\caption{Average incremental accuracy (\%) with standard deviations on CIFAR-100 and Tiny-ImageNet using a ResNet-32 backbone. Results are averaged over five runs. Baseline results are adopted from~\cite{tongmodel}.}
\label{appendix:tab_full_resnet_cl}
\centering
\resizebox{\linewidth}{!}{
\begin{tabular}{lccccccc}
\toprule
\multirow{2}{*}{Method} & \multirow{2}{*}{\parbox[c]{2cm}{\centering Model\\inversion}} & \multicolumn{3}{c}{CIFAR-100} & \multicolumn{3}{c}{Tiny-ImageNet}   \\
\cmidrule(lr){3-5}\cmidrule(lr){6-8}
                 & & 5 task      & 10 task       & 20 task    & 5 task      & 10 task       & 20 task \\
\midrule
Upper bound      & \xmark & 70.59$\pm$0.14  & 70.59$\pm$0.14  & 70.59$\pm$0.14 & 55.25$\pm$0.41 & 55.25$\pm$0.41 & 55.25$\pm$0.41 \\
\midrule
SSRE             & \xmark & 30.39$\pm$0.04 & 17.77$\pm$0.14 & 10.97$\pm$0.48 & 26.22$\pm$0.32 & 18.58$\pm$0.43 & 10.09$\pm$0.34 \\
PRAKA            & \xmark & 37.74$\pm$0.48 & 26.72$\pm$0.38 & 16.32$\pm$0.66 & 31.69$\pm$0.20 & 22.37$\pm$0.14 & 13.62$\pm$0.42 \\
\midrule
DeepInversion    & \gmark & 20.48$\pm$1.11  & 11.26$\pm$0.46  & 5.63$\pm$0.10  & - & - & - \\
ABD              & \gmark & 48.84$\pm$0.33  & 36.75$\pm$0.45  & 24.40$\pm$0.60 & 30.83$\pm$0.46 & 23.17$\pm$0.45 & 14.61$\pm$0.47 \\
DCMI             & \gmark & 41.05$\pm$0.67  & 27.70$\pm$1.07  & 18.09$\pm$0.85 & 35.78$\pm$0.37 & 25.89$\pm$0.17 & 17.03$\pm$0.08 \\
R-DFCIL          & \gmark & 49.87$\pm$0.45  & 41.80$\pm$0.24  & 31.54$\pm$0.54 & 35.33$\pm$0.02 & 29.05$\pm$0.28 & 24.85$\pm$0.16 \\
\midrule
PMI w/o CFS & \gmark & 52.05$\pm$0.02  & 43.23$\pm$0.28 & 32.23$\pm$0.42 & 37.65$\pm$0.24 & 32.09$\pm$0.20 & 25.51$\pm$0.56 \\
PMI         & \gmark & \textbf{\textcolor{blue}{52.38$\pm$0.53}} & \textbf{\textcolor{blue}{43.90$\pm$0.35}} & \textbf{\textcolor{blue}{32.60$\pm$0.29}} & \textbf{\textcolor{blue}{37.90$\pm$0.10}} & \textbf{\textcolor{blue}{32.43$\pm$0.09}} & \textbf{\textcolor{blue}{25.67$\pm$0.71}} \\
\midrule
\rowcolor{Orange!10}
\textbf{\our{}}         & \gmark & \textbf{\textcolor{red}{52.94 $\pm$ 0.07}} & \textbf{\textcolor{red}{45.38 $\pm$ 0.13}} & \textbf{\textcolor{red}{33.11 $\pm$ 0.28}} & \textbf{\textcolor{red}{38.64 $\pm$ 0.35}} & \textbf{\textcolor{red}{33.39 $\pm$ 0.11}} & \textbf{\textcolor{red}{27.11 $\pm$ 0.31}} \\
\bottomrule
\end{tabular}
}
\end{table}

\begin{figure}[t]
    \centering
    \includegraphics[width=\linewidth]{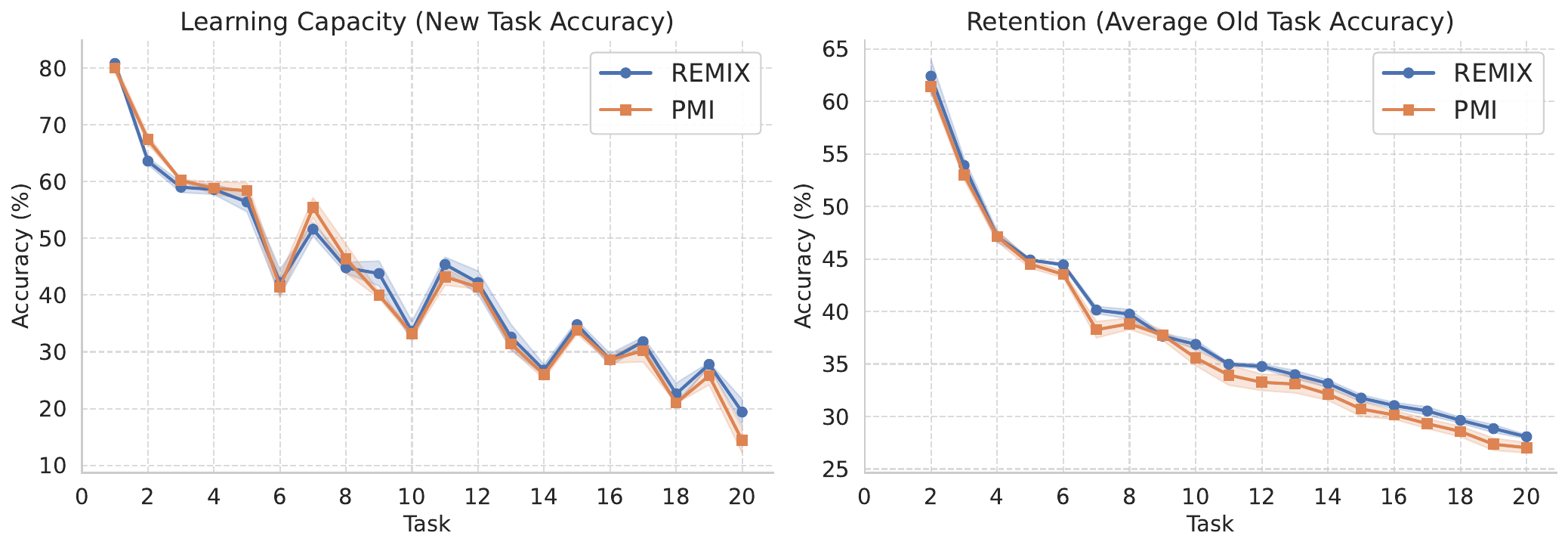}
    \caption{Task-wise performance on Tiny-ImageNet with a ResNet-32 backbone. Left: accuracy on the current task (learning capacity). Right: average accuracy on previously learned tasks (retention). Solid lines denote mean over three runs, with shaded regions indicating standard deviation. \our{} improves retention while maintaining comparable learning capacity.}
    \label{fig:resnet32_tinyimagenet_acc}
\end{figure}

\paragraph{Tiny-ImageNet.}
We further evaluate \our{} on Tiny-ImageNet using a ResNet-32 backbone. Figure~\ref{fig:resnet32_tinyimagenet_acc} presents task-wise performance, where the left plot shows accuracy on the current task (learning capacity) and the right plot shows the average accuracy on previously learned tasks (retention), both averaged over three runs.

Consistent with the observations on CIFAR-100, \our{} maintains comparable performance to PMI on newly introduced tasks, indicating that the ability to acquire new knowledge is preserved. At the same time, \our{} achieves consistently higher accuracy on previously learned tasks, demonstrating improved retention. This confirms that modeling structured feature dependencies effectively reduces forgetting without compromising learning capacity.

\subsection{Qualitative Analysis of Generated Samples}

\paragraph{Implementation Details and Architectural Adaptations.}
We generate synthetic samples via model inversion applied to pretrained backbones (ResNet-34 and ViT-B/16). Starting from random noise, input pixels are optimized using Adam with a cosine annealing schedule. We adopt a multi-resolution pyramid (e.g., $112\!\times\!112 \rightarrow 224\!\times\!224$) to progressively refine global structure. During the early stages, strong augmentations (random flips, affine transformations, spatial jitter) are applied to suppress high-frequency artifacts.

For this experiment, we adopt the DeepInversion optimization pipeline as the base inversion procedure. While PMI introduces a per-layer inversion scheme, in practice, it produces samples through a similar objective and optimization process, differing primarily in convergence speed. We therefore use DeepInversion to ensure a consistent and stable optimization setup, while isolating the effect of our contribution to feature distribution modeling. Importantly, all methods share identical optimization procedures, augmentations, and hyperparameters, and differ only in how feature statistics are modeled.

We directly optimize pixels rather than training a parametric generator; full loss definitions are provided in Appendix~\ref{appendix:sec_generator_training_details}.

\paragraph{Architectural Adaptation of LCM.}
The LCM is adapted to match the structure of each backbone.

\textbf{ResNet-34.} We extract intermediate feature maps from residual stages (\texttt{layer1}--\texttt{layer4}), obtaining multi-scale representations of shape $(B, C, H, W)$. These are flattened into high-dimensional vectors ($D = C \!\times\! H \!\times\! W$, up to $\sim\!10^6$), and LCM is fitted directly to model dependencies across both spatial and channel dimensions. During inversion, we evaluate the exact $\mathcal{O}(D\log D)$ NLL of generated features at each layer and use it as a density-matching prior, together with BatchNorm statistics alignment and classification loss.

\textbf{ViT-B/16.} Features are represented as sequences of patch tokens from transformer blocks. We discard the \texttt{[CLS]} token and retain only spatial tokens, which are reshaped into vectors for distribution modeling. LCM is applied to these representations to capture inter-patch dependencies during inversion. Since ViTs use Layer Normalization instead of BatchNorm, we match empirical means and variances computed over both batch and sequence dimensions at each block. This stabilizes global statistics while LCM enforces structured feature correlations.

The complete generation pipeline and implementation details are provided in the accompanying codebase.

\begin{figure*}[t]
\centering
\setlength{\tabcolsep}{2pt}

\begin{tabular}{c@{\;}c@{\quad\;}c}
& \textbf{Diagonal Covariance} & \textbf{Full-Feature Covariance} \\[6pt]

\rotatebox{90}{\textbf{\small \phantom{ooo}Car}} 
& 
\includegraphics[width=0.11\linewidth]{figures/vit_synthetic_samples/baseline_cars/dream_car_14.pdf}\hspace{1pt}
\includegraphics[width=0.11\linewidth]{figures/vit_synthetic_samples/baseline_cars/dream_car_24.pdf}\hspace{1pt}
\includegraphics[width=0.11\linewidth]{figures/vit_synthetic_samples/baseline_cars/dream_car_31.pdf}\hspace{1pt}
\includegraphics[width=0.11\linewidth]{figures/vit_synthetic_samples/baseline_cars/dream_car_42.pdf}
 & 
\includegraphics[width=0.11\linewidth]{figures/vit_synthetic_samples/laplace_cars/dream_car_18.pdf}\hspace{1pt}
\includegraphics[width=0.11\linewidth]{figures/vit_synthetic_samples/laplace_cars/dream_car_32.pdf}\hspace{1pt}
\includegraphics[width=0.11\linewidth]{figures/vit_synthetic_samples/laplace_cars/dream_car_34.pdf}\hspace{1pt}
\includegraphics[width=0.11\linewidth]{figures/vit_synthetic_samples/laplace_cars/dream_car_46.pdf}
\\

\rotatebox{90}{\textbf{\small \phantom{ooo}Dog}} 
& 
\includegraphics[width=0.11\linewidth]{figures/vit_synthetic_samples/baseline_dogs/dream_dog_3.pdf}\hspace{1pt}
\includegraphics[width=0.11\linewidth]{figures/vit_synthetic_samples/baseline_dogs/dream_dog_4.pdf}\hspace{1pt}
\includegraphics[width=0.11\linewidth]{figures/vit_synthetic_samples/baseline_dogs/dream_dog_28.pdf}\hspace{1pt}
\includegraphics[width=0.11\linewidth]{figures/vit_synthetic_samples/baseline_dogs/dream_dog_43.pdf}
& 
\includegraphics[width=0.11\linewidth]{figures/vit_synthetic_samples/laplace_dogs/dream_dog_12.pdf}\hspace{1pt}
\includegraphics[width=0.11\linewidth]{figures/vit_synthetic_samples/laplace_dogs/dream_dog_24.pdf}\hspace{1pt}
\includegraphics[width=0.11\linewidth]{figures/vit_synthetic_samples/laplace_dogs/dream_dog_41.pdf}\hspace{1pt}
\includegraphics[width=0.11\linewidth]{figures/vit_synthetic_samples/laplace_dogs/dream_dog_47.pdf}
\\

\rotatebox{90}{\textbf{\small \phantom{oo}Koala}} 
& 
\includegraphics[width=0.11\linewidth]{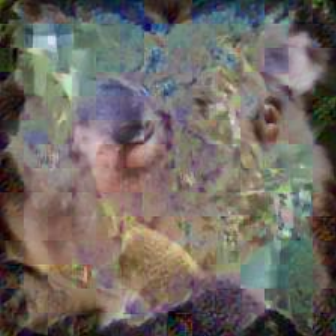}\hspace{1pt}
\includegraphics[width=0.11\linewidth]{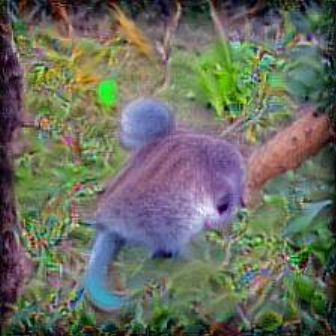}\hspace{1pt}
\includegraphics[width=0.11\linewidth]{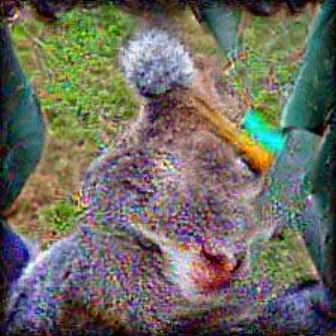}\hspace{1pt}
\includegraphics[width=0.11\linewidth]{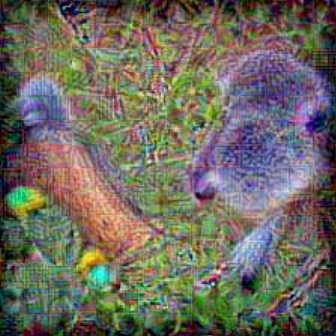}
& 
\includegraphics[width=0.11\linewidth]{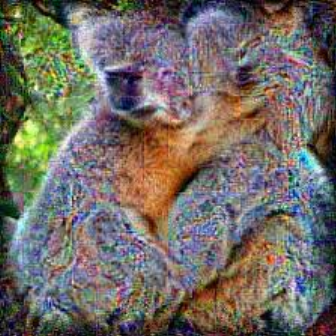}\hspace{1pt}
\includegraphics[width=0.11\linewidth]{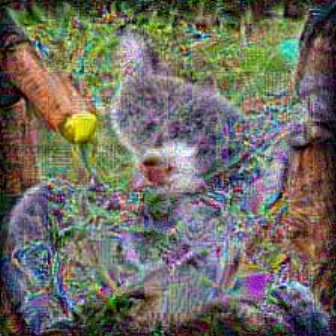}\hspace{1pt}
\includegraphics[width=0.11\linewidth]{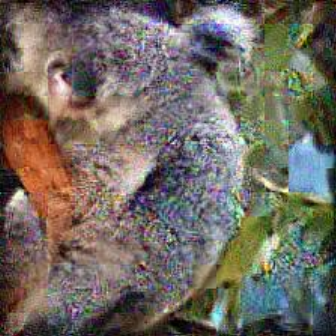}\hspace{1pt}
\includegraphics[width=0.11\linewidth]{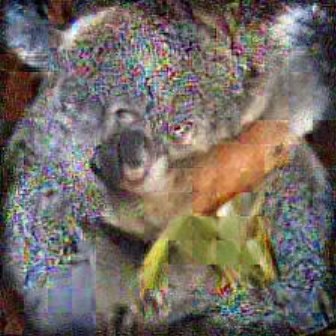}
\\

\rotatebox{90}{\textbf{\small Greenhouse}} 
& 
\includegraphics[width=0.11\linewidth]{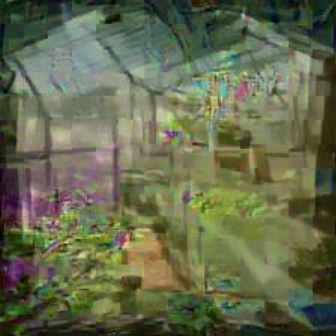}\hspace{1pt}
\includegraphics[width=0.11\linewidth]{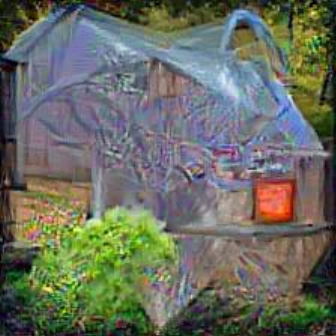}\hspace{1pt}
\includegraphics[width=0.11\linewidth]{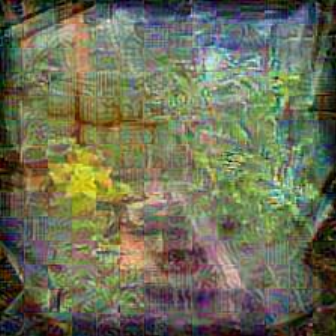}\hspace{1pt}
\includegraphics[width=0.11\linewidth]{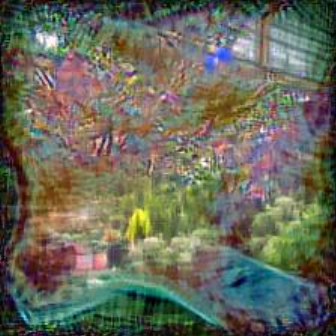}
& 
\includegraphics[width=0.11\linewidth]{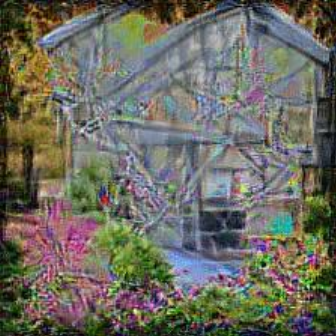}\hspace{1pt}
\includegraphics[width=0.11\linewidth]{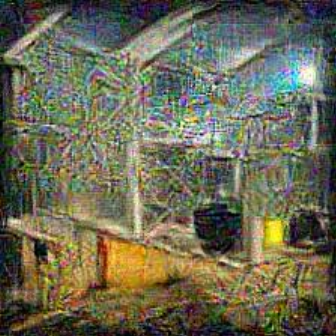}\hspace{1pt}
\includegraphics[width=0.11\linewidth]{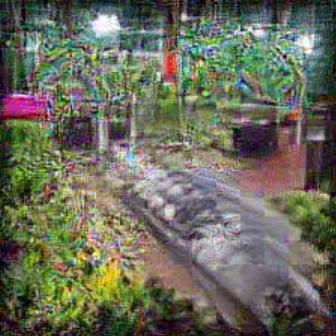}\hspace{1pt}
\includegraphics[width=0.11\linewidth]{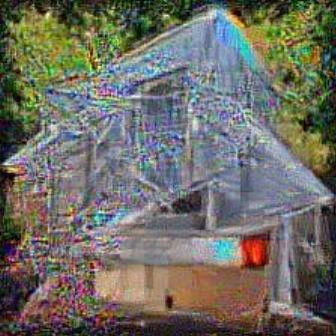}

\end{tabular}

\caption{
Qualitative comparison of generated samples using the ViT backbone. 
Left: prior approaches (e.g., PMI, DeepInversion) relying on diagonal covariance, which neglect dependencies across feature dimensions. 
Right: \our{}, which efficiently models correlations over the full feature map (up to millions of elements) while remaining computationally tractable. 
This leads to samples with improved structural coherence, fewer artifacts, and stronger semantic consistency.
}
\label{appendix:fig_vit_samples}
\end{figure*}

\begin{figure*}[t]
\centering
\setlength{\tabcolsep}{2pt}

\begin{tabular}{c@{\;}c@{\quad\;}c}
& \textbf{Diagonal Covariance} & \textbf{Full-Feature Covariance} \\

\rotatebox{90}{\textbf{\small \phantom{ooo}Car}} 
& 
\includegraphics[width=0.11\linewidth]{figures/resnet34_synthetic_samples/baseline_cars/dream_car_47.pdf}\hspace{1pt}
\includegraphics[width=0.11\linewidth]{figures/resnet34_synthetic_samples/baseline_cars/dream_car_50.pdf}\hspace{1pt}
\includegraphics[width=0.11\linewidth]{figures/resnet34_synthetic_samples/baseline_cars/dream_car_48.pdf}\hspace{1pt}
\includegraphics[width=0.11\linewidth]{figures/resnet34_synthetic_samples/baseline_cars/dream_car_49.pdf}
 & 
\includegraphics[width=0.11\linewidth]{figures/resnet34_synthetic_samples/laplace_cars/dream_car_12.pdf}\hspace{1pt}
\includegraphics[width=0.11\linewidth]{figures/resnet34_synthetic_samples/laplace_cars/dream_car_13.pdf}\hspace{1pt}
\includegraphics[width=0.11\linewidth]{figures/resnet34_synthetic_samples/laplace_cars/dream_car_24.pdf}\hspace{1pt}
\includegraphics[width=0.11\linewidth]{figures/resnet34_synthetic_samples/laplace_cars/dream_car_37.pdf}
\\

\rotatebox{90}{\textbf{\small \phantom{ooo}Dog}} & 
\includegraphics[width=0.11\linewidth]{figures/resnet34_synthetic_samples/baseline_dogs/dream_dog_27.pdf}\hspace{1pt}
\includegraphics[width=0.11\linewidth]{figures/resnet34_synthetic_samples/baseline_dogs/dream_dog_39.pdf}\hspace{1pt}
\includegraphics[width=0.11\linewidth]{figures/resnet34_synthetic_samples/baseline_dogs/dream_dog_42.pdf}\hspace{1pt}
\includegraphics[width=0.11\linewidth]{figures/resnet34_synthetic_samples/baseline_dogs/dream_dog_44.pdf}
& 
\includegraphics[width=0.11\linewidth]{figures/resnet34_synthetic_samples/laplace_dogs/dream_dog_1.pdf}\hspace{1pt}
\includegraphics[width=0.11\linewidth]{figures/resnet34_synthetic_samples/laplace_dogs/dream_dog_23.pdf}\hspace{1pt}
\includegraphics[width=0.11\linewidth]{figures/resnet34_synthetic_samples/laplace_dogs/dream_dog_13.pdf}\hspace{1pt}
\includegraphics[width=0.11\linewidth]{figures/resnet34_synthetic_samples/laplace_dogs/dream_dog_47.pdf}
\\

\rotatebox{90}{\textbf{\small \phantom{oo}Koala}} & 
\includegraphics[width=0.11\linewidth]{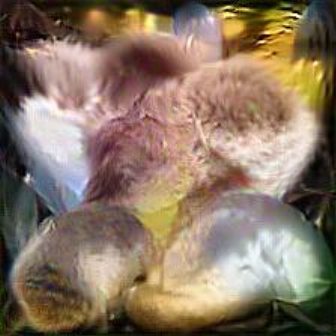}\hspace{1pt}
\includegraphics[width=0.11\linewidth]{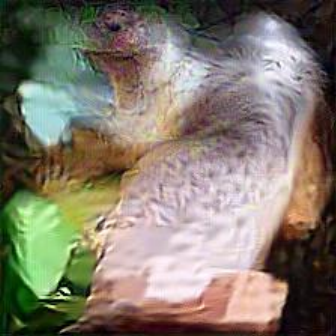}\hspace{1pt}
\includegraphics[width=0.11\linewidth]{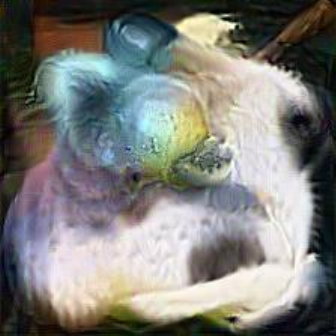}\hspace{1pt}
\includegraphics[width=0.11\linewidth]{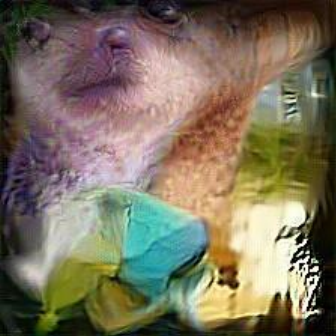}
& 
\includegraphics[width=0.11\linewidth]{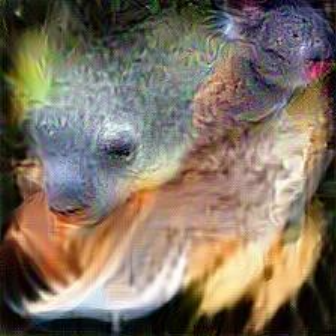}\hspace{1pt}
\includegraphics[width=0.11\linewidth]{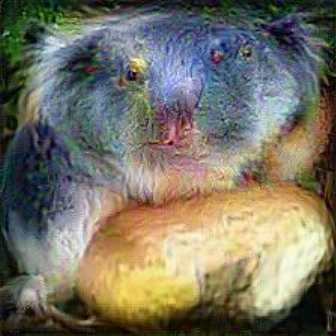}\hspace{1pt}
\includegraphics[width=0.11\linewidth]{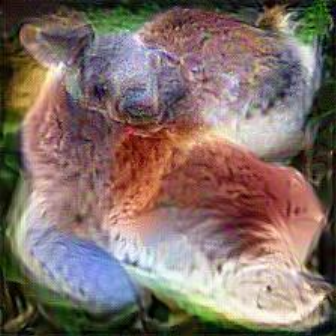}\hspace{1pt}
\includegraphics[width=0.11\linewidth]{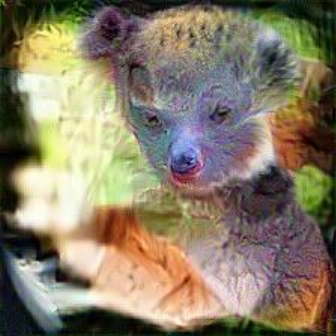}
\\

\rotatebox{90}{\textbf{\small Greenhouse}} & 
\includegraphics[width=0.11\linewidth]{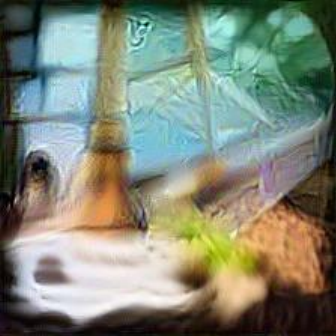}\hspace{1pt}
\includegraphics[width=0.11\linewidth]{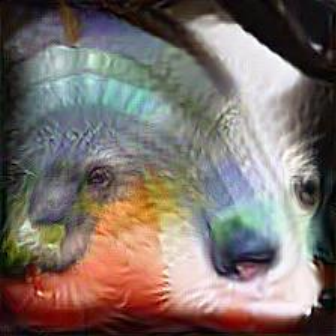}\hspace{1pt}
\includegraphics[width=0.11\linewidth]{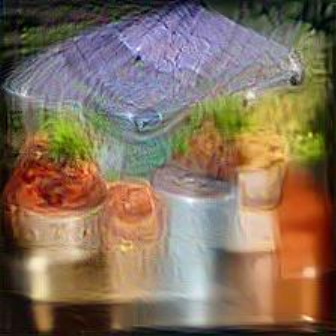}\hspace{1pt}
\includegraphics[width=0.11\linewidth]{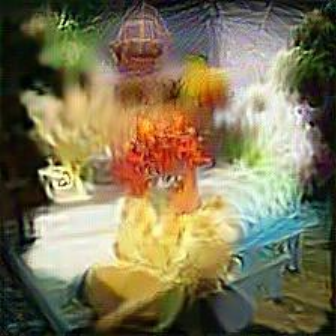}
& 
\includegraphics[width=0.11\linewidth]{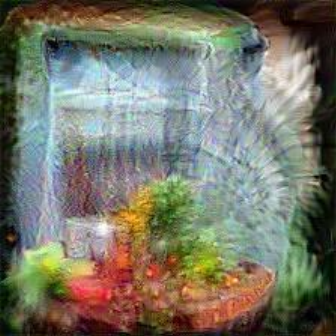}\hspace{1pt}
\includegraphics[width=0.11\linewidth]{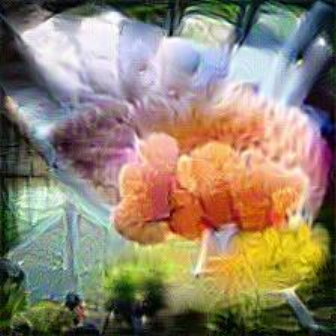}\hspace{1pt}
\includegraphics[width=0.11\linewidth]{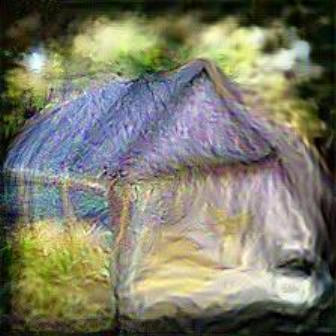}\hspace{1pt}
\includegraphics[width=0.11\linewidth]{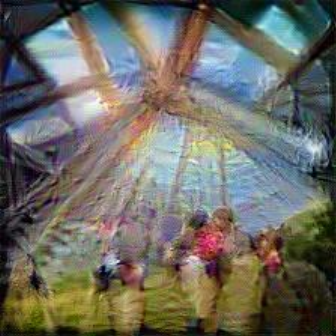}

\end{tabular}

\caption{
Qualitative comparison of generated samples using the ResNet-34 backbone. Left: prior methods based on diagonal covariance estimation, which ignore dependencies across feature dimensions.
Right: \our{}, which models correlations over the full feature map (up to millions of elements) while remaining computationally tractable.
}
\label{appendix:fig_resnet34_samples}
\end{figure*}

\paragraph{Qualitative Analysis of Generated Samples.}
Figures~\ref{appendix:fig_vit_samples} and~\ref{appendix:fig_resnet34_samples} present qualitative comparisons of synthetic samples generated using ViT-B/16 and ResNet-34 backbones, respectively. Across both architectures, methods based on diagonal covariance assumptions produce samples with weaker structural consistency, visible artifacts, and less coherent semantics. In contrast, \our{} generates samples with clearer object boundaries, more consistent textures, and stronger semantic alignment across all categories. These improvements are particularly noticeable for fine-grained categories such as \emph{koala}, where local textures and object structure are better preserved, as well as for scene-level classes such as \emph{greenhouse}, where \our{} produces more globally coherent layouts. Overall, the results suggest that modeling structured feature dependencies leads to substantially more coherent synthetic representations across both convolutional and transformer-based architectures.

\subsection{Effect of Covariance Modeling}

We analyze the representational capacity of the proposed LCM on the ViT-B/16 backbone by evaluating the log-likelihood of feature distributions across transformer blocks. Specifically, during LCM parameter fitting, we compare the total log-likelihood obtained under the proposed covariance parameterization with a diagonal Gaussian baseline trained via maximum likelihood estimation (MLE). 

To focus on spatial representations, we discard the \texttt{[CLS]} token and aggregate patch embeddings through average pooling, resulting in a $D=768$ feature vector for each image at every transformer block. Empirically, this aggregation produces more stable and accurate density estimates than directly modeling token-level representations, unlike in the ResNet-34 setting where full spatial feature maps are modeled explicitly. The reasons for this behavior are discussed below.

\begin{figure}[t]
    \centering
    \resizebox{\linewidth}{!}{\input{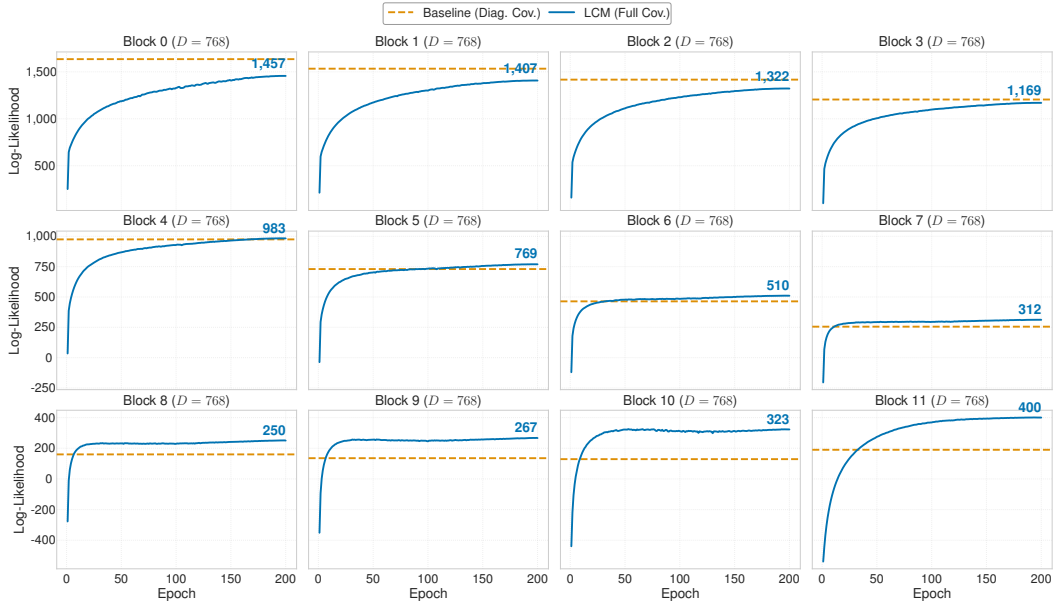}}
    \caption{Log-likelihood comparison between diagonal and full-feature covariance models across ViT-B/16 blocks. While diagonal covariance performs better in early layers with weakly correlated patch features, LCM achieves higher likelihood in deeper layers where self-attention induces structured dependencies.}
    \label{fig:vit_ll_analysis}
\end{figure}

Figure~\ref{fig:vit_ll_analysis} shows a clear dependence on representation depth. In early blocks (0--4), the diagonal model achieves higher likelihood than LCM. This aligns with the structure of ViTs, where initial layers process non-overlapping patches, producing features that are largely local and weakly correlated. In this regime, modeling full covariance offers limited benefit and can introduce unnecessary complexity.

From intermediate layers onward (around Block 5), LCM begins to outperform the diagonal baseline, with the gap increasing in deeper layers (8--11). At these stages, self-attention aggregates information across patches, yielding more structured and correlated representations. The improved likelihood indicates that LCM effectively captures these dependencies, providing a more accurate model of the feature distribution.

\subsection{Computational Efficiency}

We evaluate the computational efficiency of \our{} in comparison to PMI. As shown in Tables~\ref{tab:resnet_comparison} and \ref{tab:vit_comparison}, \our{} consistently improves performance while maintaining comparable training time and GPU memory usage across both ResNet-32 and ViT backbones.

The efficiency of our method stems from the structured LCM parameterization, which enables explicit covariance modeling while remaining computationally tractable. Naively evaluating the Gaussian NLL with a dense covariance matrix requires $\mathcal{O}(C^2)$ memory and $\mathcal{O}(C^3)$ computation due to inversion and log-determinant operations, making full-covariance modeling impractical for modern deep representations. In contrast, our formulation imposes a 1D conditional structure that enables exact likelihood evaluation in $\mathcal{O}(C\log{C})$ time.

\paragraph{Moderate-Dimensional Regime (Channel/Token Level).}
When modeling correlations over channel-wise features (ResNet) or token embeddings (ViT), the dimensionality remains moderate. In this setting, we fit the LCM to explicitly model the covariance structure of features, i.e., the underlying density of the representation space.

Although we store only the compact LCM parameters, the corresponding covariance matrix can be reconstructed when needed. As its size is manageable, it can be explicitly materialized, enabling direct optimization of covariance matching objectives. Specifically, with $\hat{\boldsymbol{\Sigma}} = \frac{1}{N} \sum_{i=1}^N \mathbf{v}_i \mathbf{v}_i^\top$, we can form $\boldsymbol{\Sigma}$ and directly minimize $\|\boldsymbol{\Sigma} - \hat{\boldsymbol{\Sigma}}\|_\mathrm{F}^2$, instead of relying on the decomposed Frobenius formulation (Eq.~\eqref{eq:frobenius_loss_function}).

Crucially, once the feature distribution is modeled, we can directly optimize the exact Gaussian NLL during model inversion. This allows synthetic samples to be generated by maximizing likelihood under the learned full-covariance model.

\paragraph{High-Dimensional Regime (Full Feature Maps).}
When extending covariance modeling to full spatial feature maps, the dimensionality grows to $D = C \times H \times W$, making naive covariance estimation completely intractable. As shown in Table~\ref{tab:resnet34_memory}, a dense covariance would require up to $199.30$ GB of memory for a ResNet-34 backbone.

In contrast, the LCM parameterization reduces this requirement to only $5.74$ MB, achieving a $35{,}541\times$ reduction. Importantly, while dense methods cannot even represent the covariance in this regime, our structured formulation allows us to both represent and use it.

In this setting, the covariance matrix is never explicitly materialized due to its size. Instead, we leverage the LCM structure to compute the exact Gaussian NLL directly in $\mathcal{O}(D\log{D})$ time (as presented in Appendix~\ref{appendix:sec_generator_training_details}). When covariance matching is required, we employ a Frobenius objective that is evaluated implicitly through the LCM using Eq.~\eqref{eq:frobenius_loss_function}, without constructing dense matrices.

\paragraph{Summary.}
Overall, \our{} makes full-covariance modeling of deep feature distributions practical. It supports explicit optimization in moderate dimensions and remains scalable in high-dimensional regimes by avoiding dense matrix representations while preserving exact likelihood computation. This enables model inversion with full-feature dependencies, which is otherwise computationally infeasible.

\begin{table}[htbp] \centering \caption{Memory and computational cost comparison for full spatial feature map covariance estimation on ResNet-34. By avoiding $\mathcal{O}(D^2)$ dense matrix allocations, our LCM formulation reduces the total memory footprint from $199.30$ GB to $5.74$ MB, achieving a $35{,}541\times$ efficiency gain.} \begin{tabular}{lcccc} \toprule Layer & Dimension ($D$) & LCM Mem. & Dense Mem. & Savings \\ \midrule Layer 1 & 200,704 & \textbf{3.06 MB} & 150.06 GB & 50,176$\times$ \\ Layer 2 & 100,352 & \textbf{1.53} MB & 37.52 GB & 25,088$\times$ \\ Layer 3 & 50,176 & \textbf{0.77 MB} & 9.38 GB & 12,544$\times$ \\ Layer 4 & 25,088 & \textbf{0.38 MB} & 2.34 GB & 6,272$\times$ \\ \midrule Total & 376,320 & \textbf{5.74 MB} & 199.30 GB & 35,541$\times$ \\ \bottomrule \end{tabular} \label{tab:resnet34_memory} \end{table} 

\begin{table}[htbp] \centering \caption{Comparison with PMI on CIFAR-100 (20 tasks, 5 classes per task) using a ResNet-32 backbone. \our{} achieves higher final accuracy with comparable time and memory usage.} \begin{tabular}{lccc} \toprule Method & Last & Time & Total GPU Mem. (GB) \\ \midrule PMI & 52.38$\pm$0.53 & \textbf{1h 27m} & \textbf{7.28} \\ \our{} & \textbf{52.94$\pm$0.07} & 1h 37m & \textbf{7.28} \\ \bottomrule \end{tabular} \label{tab:resnet_comparison} \end{table} \begin{table}[htbp] \centering \caption{Comparison with PMI on CIFAR-100 (10 tasks, 10 classes per task) using a ViT backbone with MoE-Adapter. \our{} improves performance with negligible computational overhead.} \begin{tabular}{lccc} \toprule Method & Avg. & Time & Total GPU Mem. (GB) \\ \midrule PMI & 88.35 & \textbf{3h 33m} & \textbf{7.39} \\ \our{} & \textbf{88.47} & 3h 42m & \textbf{7.39} \\ \bottomrule \end{tabular} \label{tab:vit_comparison} \end{table}



\end{document}